\documentclass[lettersize,journal]{IEEEtran}
\usepackage{amsmath,amsfonts}
\usepackage{algorithmic}
\usepackage{algorithm}
\usepackage{array}
\usepackage[caption=false,font=normalsize,labelfont=sf,textfont=sf]{subfig}
\usepackage{textcomp}
\usepackage{stfloats}
\usepackage{url}
\usepackage{verbatim}
\usepackage{graphicx}
\usepackage{cite}
\usepackage{color,colortbl}
\definecolor{blush}{rgb}{0.87, 0.36, 0.51}
\usepackage[colorlinks,citecolor=green]{hyperref}

\usepackage{amsmath}
\usepackage{amssymb}
\definecolor{bblue}{rgb}{0.36, 0.51, 0.87}
\definecolor{ggray}{rgb}{0.88, 0.87, 0.87}
\newcolumntype{a}{>{\columncolor{ggray}}c}

\usepackage[capitalize]{cleveref}
\crefname{section}{Sec.}{Secs.}
\Crefname{section}{Section}{Sections}
\Crefname{table}{Table}{Tables}
\crefname{table}{Tab.}{Tabs.}

\usepackage{bbm}

\usepackage{forest}

\usepackage{amsthm}
\usepackage{cancel}
\usepackage{caption}
\theoremstyle{definition}



\newcommand{\stare}{\faStar[regular]}
\newcommand{\starh}{\faStarHalf*}
\newcommand{\starf}{\faStar}

\makeatletter
\newcommand*\bigcdot{\mathpalette\bigcdot@{.5}}
\newcommand*\bigcdot@[2]{\mathbin{\vcenter{\hbox{\scalebox{#2}{$\m@th#1\bullet$}}}}}
\makeatother

\usepackage{makecell}
\usepackage{ragged2e}
\usepackage{enumitem}
\usepackage{tikz}
\usepackage{fontawesome5}
\usetikzlibrary{shapes.geometric, arrows, positioning, fit}
\usepackage{multirow}
\usepackage{booktabs}

\begin{document}

\title{Adapting Vision-Language Models Without Labels: A Comprehensive Survey}

\author{Hao Dong$^*$, Lijun Sheng$^*$, Jian Liang$^\dag$, Ran He, Eleni Chatzi, Olga Fink
\thanks{$^*$ Equal contribution. $^\dag$ Corresponding author.}
\thanks{H.~Dong and E.~Chatzi are with ETH Z\"urich, Switzerland. (Email: hao.dong@ibk.baug.ethz.ch; chatzi@ibk.baug.ethz.ch).}
\thanks{L.~Sheng is with the University of Science and Technology of China. (Email: slj0728@mail.ustc.edu.cn).}
\thanks{J.~Liang and R.~He are with NLPR \& MAIS, Institute of Automation, Chinese Academy of Sciences. (Email: liangjian92@gmail.com; rhe@nlpr.ia.ac.cn).}
\thanks{O.~Fink is with EPFL, Switzerland. (Email: olga.fink@epfl.ch).}
\thanks{Manuscript received \today.}
}

\markboth{Journal of \LaTeX\ Class Files,~Vol.~14, No.~8, August~2021}%
{Shell \MakeLowercase{\textit{et al.}}: A Sample Article Using IEEEtran.cls for IEEE Journals}


\maketitle

\begin{abstract}
Vision-Language Models (VLMs) have demonstrated remarkable generalization capabilities across a wide range of tasks. However, their performance often remains suboptimal when directly applied to specific downstream scenarios without task-specific adaptation. To enhance their utility while preserving data efficiency, recent research has increasingly focused on unsupervised adaptation methods that do not rely on labeled data. Despite the growing interest in this area, there remains a lack of a unified, task-oriented survey dedicated to unsupervised VLM adaptation.
To bridge this gap, we present a comprehensive and structured overview of the field. We propose a taxonomy based on the availability and nature of unlabeled visual data, categorizing existing approaches into four key paradigms: \textit{Data-Free Transfer} (no data), \textit{Unsupervised Domain Transfer} (abundant data), \textit{Episodic Test-Time Adaptation} (batch data), and \textit{Online Test-Time Adaptation} (streaming data).
Within this framework, we analyze core methodologies and adaptation strategies associated with each paradigm, aiming to establish a systematic understanding of the field. Additionally, we review representative benchmarks across diverse applications and highlight open challenges and promising directions for future research.
An actively maintained repository of relevant literature is available at \url{https://github.com/tim-learn/Awesome-LabelFree-VLMs}.
\end{abstract}


\ifCLASSOPTIONcompsoc
\IEEEraisesectionheading{\section{Introduction}\label{sec:introduction}}
\else
\section{Introduction}
\label{sec:introduction}
\fi

\IEEEPARstart{V}{ision}-language models (VLMs), such as CLIP~\cite{radford2021learning}, ALIGN~\cite{jia2021scaling}, Flamingo~\cite{alayrac2022flamingo}, and LLaVA~\cite{liu2023visual} have attracted considerable attention from both academia and industry due to their powerful cross-modal reasoning capabilities.
These models learn joint image-text representations from large-scale datasets~\cite{schuhmann2022laion} and have demonstrated impressive zero-shot performance and generalization across a variety of tasks.
VLMs have been successfully applied in diverse domains, including autonomous driving~\cite{chen2023clip2scene}, robotics~\cite{shafiullah2022clip}, anomaly detection~\cite{sun2025ano}, and cross-modal retrieval~\cite{chun2021probabilistic}.

However, because the pre-training phase cannot capture the full diversity of downstream tasks and environments, adapting VLMs to specific applications remains a fundamental challenge.
Early efforts primarily relied on supervised fine-tuning \cite{zhou2022learning, zhou2022conditional, khattak2022maple, wang2024hard}, which explores more knowledge in annotated examples. 
Despite their effectiveness, they still suffer from high annotation costs and performance degradation under distribution shifts \cite{saenko2010adapting} between training and test data.
To address these limitations, a growing body of work has explored unsupervised adaptation techniques \cite{menon2023visual, zhou2022extract, huang2022unsupervised, shu2022test, ma2024swapprompt, sheng2025illusion}.
These approaches—often referred to as zero-shot inference \cite{pratt2023does, kahana2022improving, qian2024online}, test-time methods \cite{shu2022test, abdul2023align, karmanov2024efficient}, or unsupervised tuning \cite{huang2022unsupervised, li2023masked, liang2024realistic}—aim to improve VLMs' performance in downstream tasks without relying on costly annotation. 
Such methods have proven effective across a wide range of applications, including image classification \cite{huang2022unsupervised, shu2022test, menon2023visual}, segmentation \cite{zhou2022extract, shin2022reco, hajimiri2025pay}, medical image diagnosis \cite{rahman2025can, liu2023chatgpt}, and action recognition \cite{bosetti2024text, yan2024dts}.

\begin{figure*}[t]
  \centering
    \includegraphics[width=0.95\linewidth]{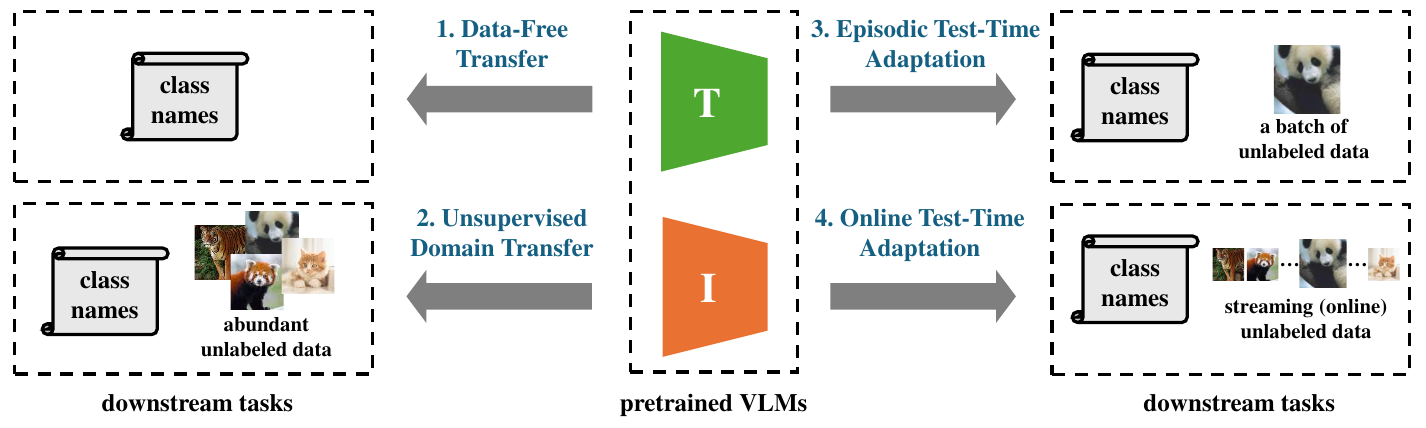}
   \vspace{-0.2cm}
   \caption{Illustration of our taxonomy on unsupervised adaptation with VLMs. We categorize existing unsupervised methods into four task paradigms based on the availability of unlabeled visual data.}
   \label{fig:taxonomy}
\end{figure*}

Given the rapid growth of this research area, this survey provides a comprehensive and structured overview of existing unsupervised adaptation methods for VLMs.
To the best of our knowledge, we are the first to introduce a taxonomy centered on the availability of unlabeled visual data—an often overlooked yet practically critical factor in real-world deployment.
As illustrated in Fig.~\ref{fig:taxonomy}, we categorize existing approaches into four paradigms:
(1) Data-Free Transfer \cite{zhou2022extract, menon2023visual, pratt2023does}, which adapts models using only textual class names;
(2) Unsupervised Domain Transfer \cite{huang2022unsupervised, tanwisuth2023pouf, kalantidis2024label}, which utilizes abundant unlabeled data from the downstream tasks;
(3) Episodic Test-Time Adaptation \cite{shu2022test, abdul2023align, zanella2024test}, which adapts models to a batch of test instances;
and (4) Online Test-Time Adaptation \cite{ma2024swapprompt, karmanov2024efficient, qian2024online}, which addresses the challenge of streaming test data.
This taxonomy provides a principled framework for understanding the landscape of unsupervised VLM adaptation, guiding practitioners in selecting suitable techniques.
We also believe our taxonomy will facilitate fair comparisons across future work within the same paradigm.

\begin{figure*}[t]
\centering
\includegraphics[width=0.95\linewidth]{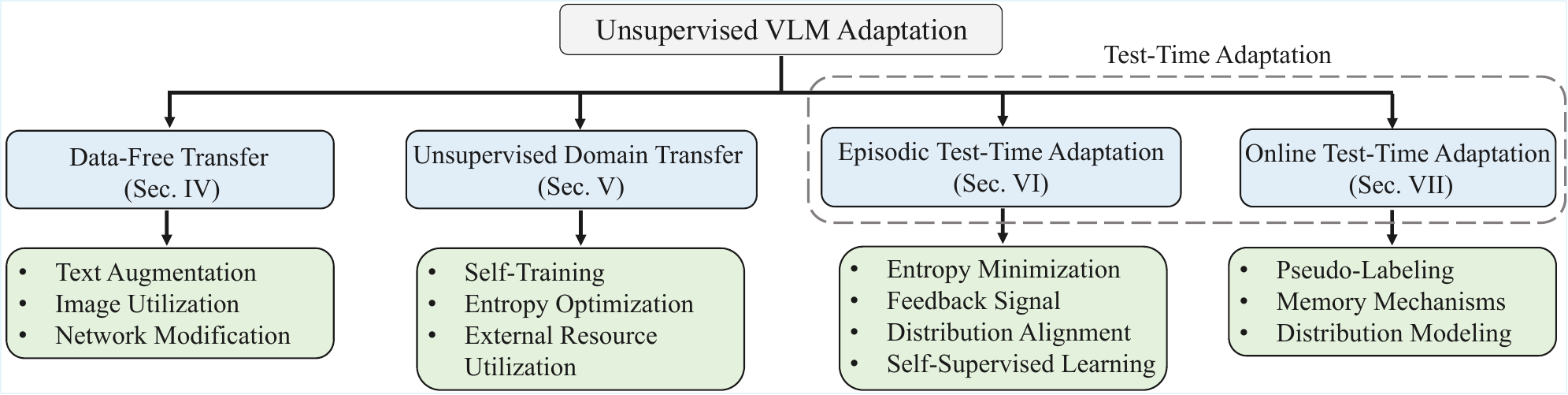}
\caption{Taxonomy of unsupervised adaptation paradigms for vision-language models (VLMs).}
\label{fig-over}
\end{figure*}

The organization of this survey follows the structure shown in~\cref{fig-over}.
Sec.~\ref{sec:related} provides an overview of several research topics related to unsupervised learning in the context of VLMs.
Sec.~\ref{sec:pre} introduces zero-shot inference with VLMs and presents a comprehensive taxonomy based on the availability of unlabeled visual data.
The central focus of this survey is discussed in Sec.~\ref{sec:datafree} - Sec.~\ref{sec:online}, where we analyze existing approaches within data-free transfer, unsupervised domain transfer, episodic test-time adaptation, and online test-time adaptation, respectively.
Sec.~\ref{sec:application} explores a variety of application scenarios that utilize unsupervised techniques and introduces related benchmarks, offering a broader perspective on their practical implications and real-world utility.
Finally, we summarize emerging trends in the field and identify key scientific questions that could inspire future work in Sec.~\ref{sec:future}.

\noindent\textbf{Comparison with previous surveys.} 
In recent years, several surveys \cite{zhang2024vision,dong2025mmdasurvey, li2025generalizing,liang2024comprehensive} have explored various aspects of unsupervised adaptation and fine-tuning of VLMs.
Existing works~\cite{liang2024comprehensive,wang2024search, xiao2024beyond} predominantly focus on unimodal model transfer, providing a thorough analysis of this domain, but they offer limited coverage of VLMs.
An early work~\cite{zhang2024vision} discusses the pre-training stage of VLMs  and briefly analyzes its fine-tuning method for vision tasks.
Another survey~\cite{dong2025mmdasurvey} discusses the adaptation and generalization of multimodal models, but at a relatively coarse-grained level.
A recent work~\cite{li2025generalizing} uses generalization to understand VLMs' downstream tasks and reviews existing methods with a perspective of the parameter space.
While these surveys contribute valuable insights, our work distinguishes itself by introducing, for the first time, a taxonomy based on the availability of unlabeled visual data and analyzing cutting-edge technologies in each of these paradigms.
We believe this is a novel and crucial contribution to the field, especially in terms of the deployment of VLMs.

\section{Related Research Topics}
\label{sec:related}

\subsection{Vision-Language Models}
Recent progress in VLMs has been remarkable, driven by the integration of large-scale pre-training~\cite{radford2021learning,yao2021filip}, transformer architectures~\cite{vaswani2017attention,dosovitskiy2021image}, and massive multimodal datasets~\cite{schuhmann2022laion,chen2015microsoft}. Models such as CLIP~\cite{radford2021learning}, ALIGN~\cite{jia2021scaling}, and Flamingo~\cite{alayrac2022flamingo} have pushed the boundaries by learning robust joint representations that bridge the semantic gap between vision and language. These advancements have enabled impressive performance across a range of tasks, including image captioning~\cite{you2016image}, visual question answering~\cite{wu2017visual}, text-to-image synthesis~\cite{rombach2022high}, and cross-modal retrieval~\cite{zhen2019deep}, often exhibiting strong zero-shot and few-shot learning capabilities. For further information on vision-language models, we refer the reader to the recent survey papers~\cite{zhang2024vision,zhou2024comprehensive}.

\subsection{Zero-Shot Learning}
Zero-shot learning (ZSL) aims to recognize unseen classes by leveraging semantic information such as attributes or word embeddings. Early methods relied on learning compatibility functions between visual features and manually defined attributes~\cite{lampert2009learning,farhadi2009describing}. Subsequent works introduced embedding-based approaches that align visual and semantic spaces using supervised objectives~\cite{frome2013devise,romera2015embarrassingly}. To address limitations in generalization, generative models were employed to synthesize visual features for unseen classes~\cite{verma2018generalized,felix2018multi,xian2019f}. Recent research emphasizes generalized settings~\cite{chao2016empirical,xian2017zero}, aiming to improve robustness and fair evaluation across seen and unseen categories.
ZSL serves as a foundational principle for unsupervised VLM adaptation, leveraging semantic descriptions (e.g., text prompts) to bridge seen and unseen classes. This enables models to generalize to novel visual concepts without requiring any labeled examples.
For further information on ZSL, we refer the reader to the recent survey papers~\cite{pourpanah2022review,wang2019survey}.

\subsection{Supervised Fine-Tuning of VLMs}
Supervised fine-tuning of VLMs has emerged as a key strategy to adapt pre-trained models to downstream tasks with task-specific supervision. Researchers have increasingly focused on parameter-efficient fine-tuning techniques—such as prompt tuning~\cite{ma2022understanding,khattak2022maple,zhou2022learning}, adapter modules~\cite{gao2024clip,zhang2021tip}, and lightweight task-specific layers~\cite{zanella2024low}—that allow models to adapt to new domains while preserving the generality of their pre-trained features. Moreover, some approaches use large language models (LLMs) to aid in adapting VLMs~\cite{pratt2023does,menon2022visual} or adapt VLMs to dense prediction tasks~\cite{rao2022denseclip,zhou2022extract}. For further information on supervised fine-tuning of VLMs, we refer the reader to the recent survey papers~\cite{dong2025mmdasurvey,gu2023systematic}. Instead of relying on explicit label supervision, this survey focuses on unsupervised VLM adaptation, where models must adapt to downstream tasks without access to annotated data.

\subsection{Source-Free Domain Adaptation} 
Source-free domain adaptation (SFDA) addresses the practical setting where access to source data is restricted during adaptation~\cite{nejjar2024sfosda}. SHOT~\cite{liang2020we} initiates this paradigm by aligning target features through information maximization and self-supervised learning. The following works improve class-wise feature structure via neighborhood clustering~\cite{wen2019exploiting} and contrastive learning~\cite {chen2022contrastive}. Other approaches leverage prototype refinement~\cite{kundu2020universal}, self-training~\cite{huang2021model}, and adversarial learning~\cite{kim2021domain} to enhance robustness. SFDA is closely related to unsupervised VLM adaptation, as the original source data used to pre-train VLMs is often massive and impractical to access during adaptation. For further information on SFDA, we refer the reader to the recent survey papers~\cite{li2024comprehensive,fang2024source}.

\subsection{Traditional Test-Time Adaptation}  
Test-time adaptation (TTA) focuses on adapting a pre-trained source model online to address distribution shifts without access to source data or target labels. Online TTA methods~\cite{wang2021tent,dong2025aeo} update specific model parameters using incoming test samples, leveraging unsupervised objectives such as entropy minimization and pseudo-labeling. Robust TTA methods~\cite{niu2022efficient,gong2023sotta} tackle challenging real-world scenarios, including label shifts, single-sample adaptation, and mixed domain shifts. Meanwhile, continual TTA approaches~\cite{wang2022continual,song2023ecotta} handle evolving distribution shifts encountered over time, which is particularly relevant in dynamic real-world applications.
Although most traditional TTA methods were introduced for vision-only architectures, their core mechanisms, such as entropy minimization and pseudo-labeling, have been repurposed for TTA of VLMs~\cite{shu2022test,feng2023diverse,abdul2023align}. 
For a comprehensive review of test-time adaptation, we refer readers to the recent survey papers~\cite{liang2024comprehensive,wang2024search}.

\section{Preliminaries}
\label{sec:pre}

\noindent\textbf{Vision-Language Models (VLMs)} typically consist of an image encoder that maps high-dimensional images into a low-dimensional embedding space and a text encoder that generates text representations from natural language. Since the introduction of CLIP~\cite{radford2021learning}, numerous improved models have been proposed, including ALIGN~\cite{jia2021scaling}, EVA-CLIP~\cite{sun2023eva}, and SigLIP~\cite{zhai2023sigmoid}, with CLIP remaining the most widely used model in existing works. CLIP is trained on $400$ million image-text pairs and aligns image and text embeddings using contrastive loss. Given a batch of image-text pairs, CLIP maximizes the cosine similarity for matched pairs while minimizing it for unmatched ones. During inference, the class names of a target dataset are embedded using the text encoder with a prompt of the form ``a photo of a [CLASS]``, where [CLASS] is replaced with specific class names (e.g., cat, dog, car). The text encoder then generates text embeddings $\mathbf{t}_c$ for each class $c$, and the prediction probability for an input image $\mathbf{x}$ with embedding $\mathbf{f}_{\mathbf{x}}$ is computed as:
\begin{equation} \label{eqn:clip-metric}
    p(y|\mathbf{x}) = \frac{\exp{(\cos \left(\mathbf{f}_\mathbf{x}, \mathbf{t}_{y} \right) / \tau )} } { \sum_{c=1}^{C} \exp{(\cos \left(\mathbf{f}_\mathbf{x}, \mathbf{t}_{c} \right) / \tau )} },
\end{equation}
where $\cos(\cdot,\cdot)$ measures the cosine similarity and $\tau$ is a temperature parameter.

\noindent\textbf{Prompt Tuning.} Instead of relying on manually crafted prompts, prompt tuning methods optimize prompts to improve performance on downstream tasks. Specifically, prompt tuning learns a prompt $\mathbf{p} = [V]_1[V]_2...[V]_M \in \mathbb{R}^{M\times d}$ in the text embedding space, where $M$ is the number of tokens and $d$ is the embedding size. Given training data $\mathcal{D}_{\text{train}} = \{(\mathbf{x}_i, y_i)\}$ from the downstream task, the objective is to generate text inputs of the form ``$[V]_1$$[V]_2$...$[V]_M$[CLASS]`` that provide the model with the most relevant context information. For image classification with cross-entropy loss $CE$, this optimization can be formulated as:
\begin{equation}
    \label{eq:p-tuning}
    \mathbf{p}^{\ast} = \text{arg} \min_{\mathbf{p}}\mathbb{E}_{(\mathbf{x}, y)\sim\mathcal{D_{\text{train}}}}CE(p(y|\mathbf{x}), y). 
\end{equation}

\noindent\textbf{Taxonomy.} In this survey, we introduce a taxonomy that systematically categorizes unsupervised VLM adaptation methods based on the availability of unlabeled visual data during the adaptation process (\cref{fig:taxonomy}). The proposed framework defines four distinct adaptation paradigms, each with unique assumptions and challenges. The first, data-free transfer, represents the most constrained setting where no visual data from the downstream task is available. In contrast, unsupervised domain transfer assumes access to an abundant, static collection of unlabeled target data, enabling a more comprehensive, offline adaptation before inference. The final two categories address adaptation that occurs during the testing phase itself. Episodic test-time adaptation operates on a small batch of test instances, adapting the model specifically for that batch. Lastly, online test-time adaptation tackles the most dynamic scenario, where the model must continuously learn from a sequential stream of incoming data points and update itself in real-time. This taxonomy highlights the fundamental differences in data access, computational constraints, and algorithmic design across the spectrum of unsupervised VLM adaptation.
In the following sections, we provide a detailed overview of existing approaches within each paradigm.


\section{Data-Free Transfer}
\label{sec:datafree}

\begin{figure*}[tbp]
\centering
\begin{minipage}[t]{0.32\textwidth}
  \centering
  \includegraphics[width=\linewidth]{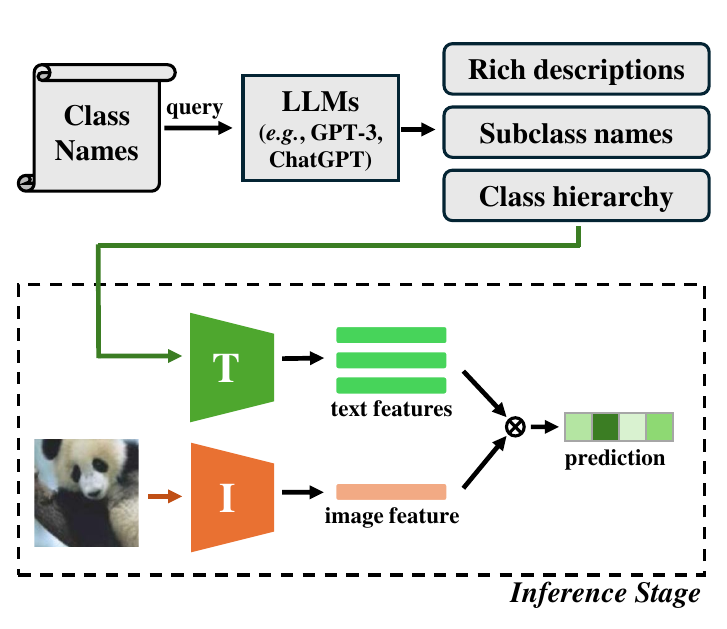}
  \vspace{-0.5cm}
  \caption*{(a) Text augmentation.}
\end{minipage}
\hfill
\begin{minipage}[t]{0.32\textwidth}
  \centering
  \includegraphics[width=\linewidth]{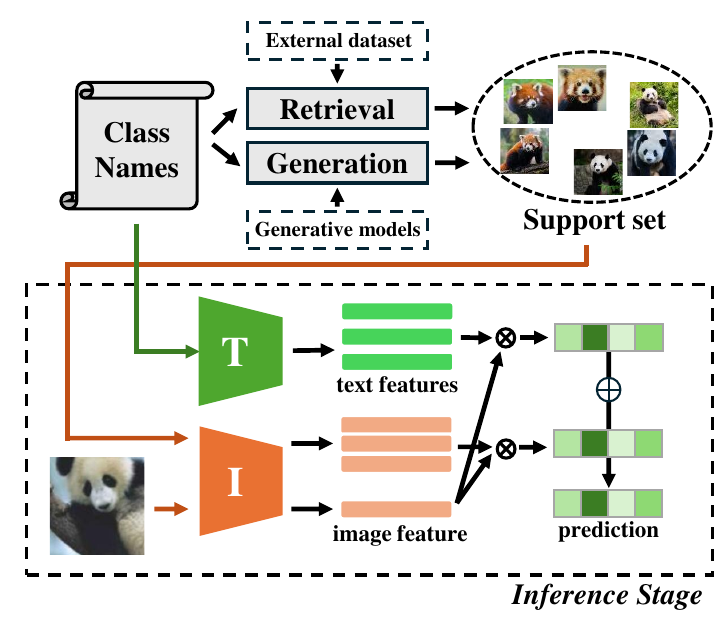}
  \vspace{-0.5cm}
  \caption*{(b) Image utilization.}
\end{minipage}
\hfill
\begin{minipage}[t]{0.32\textwidth}
  \centering
  \includegraphics[width=\linewidth]{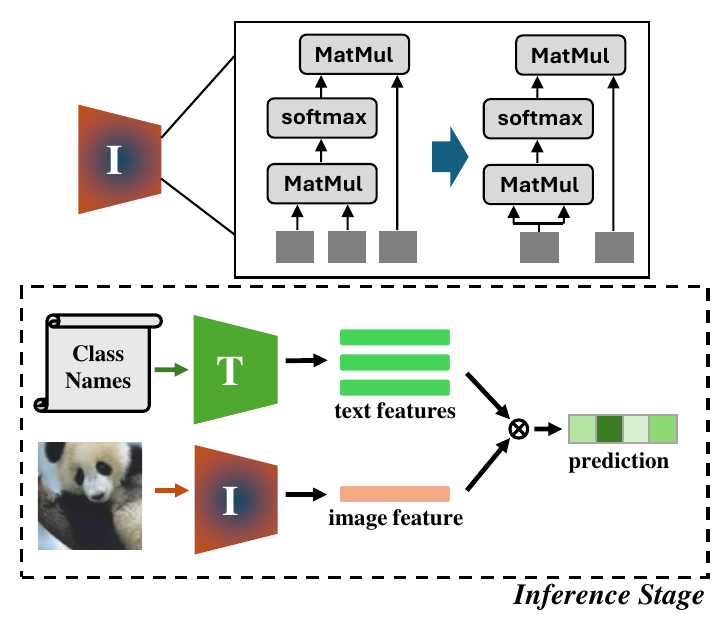}
  \vspace{-0.5cm}
  \caption*{(c) Network modification.}
\end{minipage}
   \vspace{-0.1cm}
\caption{Three representative strategies of the \textbf{data-free transfer} paradigm.}
\label{fig:data-free-methods}
\end{figure*}

\textbf{Paradigm description.}
Data-free transfer in the context of VLMs refers to adapting pre-trained models to downstream tasks \emph{without access to any visual data (e.g., images) from the downstream task}.
This setting is particularly challenging, as it relies exclusively on textual category names to guide the adaptation process.
As such, data-free transfer is considered the most difficult paradigm within unsupervised VLM adaptation.
Despite these difficulties, methods developed for this setting are often highly generalizable and broadly applicable, offering robust solutions for a variety of unsupervised tasks across domains where visual data is scarce, sensitive, or unavailable.

We review existing data-free transfer methods and categorize their strategies into three primary approaches: text augmentation, image utilization, and network modification.
These categories are summarized in Table~\ref{tab:datafree}, and we introduce each strategy in detail along with related methods in the following subsections.

\begin{table}[tbp]
\centering
\caption{Popular strategies along with their representative works of \textbf{data-free transfer}.}
\label{tab:datafree}
\resizebox{0.85\linewidth}{!}{
\begin{tabular}{ll}
\toprule
\textbf{Strategies} & \textbf{Representative Works} \\
\midrule

\multirow{2}{*}{\textbf{Text Augmentation}} & DCLIP~\cite{menon2023visual}, CuPL~\cite{pratt2023does},\\
& CHiLS~\cite{novack2023chils}, TaI~\cite{guo2023texts}. \\
\midrule
\multirow{2}{*}{\textbf{Image Utilization}} & ReCo~\cite{shin2022reco}, SuS-X~\cite{udandarao2023sus}, \\ 
& Priming~\cite{wallingford2023neural}, GenCL~\cite{seo2024just}. \\
\midrule
\multirow{2}{*}{\textbf{Network Modification}} & MaskCLIP~\cite{zhou2022extract}, CALIP~\cite{guo2023calip},   \\
& SCLIP~\cite{wang2024sclip}, ProxyCLIP~\cite{lan2024proxyclip}. \\
\bottomrule
\end{tabular}
}
\end{table}

\subsection{Text Augmentation}
In data-free transfer paradigm, where only class names are available, direct inference results in the rich semantic capacity of the text encoder remaining underexploited.
To mitigate this limitation, several methods have been proposed that enhance the textual input through text augmentation, aiming to generate more informative representations, as shown in Fig.~\ref{fig:data-free-methods} (a).
These augmented texts help unlock the latent knowledge of the text encoder, thereby improving model performance despite the absence of visual data.

Leveraging the powerful capabilities of LLMs, such as GPT-3 \cite{brown2020language}, several data-free transfer methods \cite{menon2023visual, pratt2023does, mirza2024meta} have adopted text augmentation strategies to enrich class representations with more informative and discriminative descriptions.
These approaches aim to replace simple class names with richer textual content, thereby improving alignment with visual concepts.
For instance, DCLIP \cite{menon2023visual} and CuPL \cite{pratt2023does} use GPT-3 to generate multiple semantic descriptors and full descriptive sentences, injecting discriminative knowledge into category representations.
REAL-Prompt \cite{parashar2024neglected} identifies performance drops for categories with low-frequency terms in the pretraining corpus and addresses this by prompting ChatGPT \cite{openai2023chatgpt} to substitute them with higher-frequency synonyms that are more familiar to the encoder.
MPVR \cite{mirza2024meta} introduces a two-step prompting mechanism, where LLMs first generate task-relevant queries that are then used to derive category-specific prompts, improving both relevance and diversity.
In domain-specific applications like medical image diagnosis, ChatGPT \cite{openai2023chatgpt} is used to generate symptom-based descriptions of disease classes \cite{liu2023chatgpt}.
Moreover, Parashar et al. \cite{parashar2023prompting} show that replacing scientific species names with common English terms improves classification performance.
Interestingly, reliance on LLMs is not strictly necessary.
WaffleCLIP \cite{roth2023waffling} demonstrates that even random word augmentations to class names can yield results comparable to those generated with LLMs.
In a complementary direction, TAG \cite{liu2024tag} proposes an out-of-distribution (OOD) detection approach that leverages a scoring mechanism based on permuted prompt templates. This score captures consistency of model predictions across varied phrasings and enables more accurate detection.

Extending this line of work, existing studies \cite{novack2023chils, sun2024training, bosetti2024text, moayeri2024embracing} have also explored the use of subclass names as an alternative to rich textual descriptions, aiming to describe the semantic scope of each category more precisely.
CHiLS \cite{novack2023chils} leverages GPT-3 \cite{brown2020language} to generate subclass names for each original class and makes final predictions by aggregating similarities between the image and both the superclass and its associated subclasses.
In the context of semantic segmentation, subclass prompts allow for finer-grained,  patch-level alignment with the target superclass,  leading to measurable improvements in performance \cite{sun2024training}.
For video-based action recognition, TEAR \cite{bosetti2024text} decomposes complex actions into multiple sub-actions, generating concise descriptions for each and forming a robust composite representation by averaging their features.
Beyond subclassing, other approaches enrich semantic understanding through category-related attributes, helping to capture nuanced intra-class variation and enhance image recognition accuracy  \cite{moayeri2024embracing}.
Furthermore, EOE \cite{cao2024envisioning} extends this strategy to OOD detection by prompting LLMs to generate potential OOD category names, thereby broadening the model’s recognition capacity beyond the training distribution.

Rather than classifying objects across all possible categories, some approaches \cite{ren2023chatgpt, lee2025enhancing, liang2025making} simplify the task by organizing classes into clusters or hierarchical structures, decomposing a complex classification problem into a sequence of hierarchical sub-tasks.
An early work \cite{ren2023chatgpt} constructs hierarchical clusters of candidate categories and employs ChatGPT \cite{openai2023chatgpt} to generate group-specific, discriminative textual descriptions.
More recently, Lee et al. \cite{lee2025enhancing} leverage textual feature similarity to identify semantically similar classes and then prompt an LLM to generate visual descriptors that distinguish a class from its nearest semantic neighbors.
Moving beyond accuracy alone, HAPrompts \cite{liang2025making} introduces a hierarchical classification framework to encourage the model to promote better mistakes, encouraging the model to predict semantically related labels when misclassifications occur.

Another line of data-free transfer methods \cite{guo2023texts, mirza2023tap, khattak2025learning} leverage external textual data as a training signal to guide models toward more robust task performance.
For example, TaI \cite{guo2023texts} replaces annotated images with rich textual descriptions for prompt tuning, introducing dual-grained prompts that capture both global semantic context and local discriminative features.
In a related approach, TAP \cite{mirza2023tap} constructs class-specific textual descriptions and trains a text-only classifier using cross-entropy loss. This classifier is then integrated with a visual encoder at inference time to enhance recognition accuracy.
Going a step further, ProText \cite{khattak2025learning} optimizes deep prompt parameters to steer the text encoder toward extracting meaningful representations from LLM-generated descriptions, which embed extensive linguistic knowledge and fine-grained conceptual distinctions.

\subsection{Image Utilization}
In the absence of visual data, methods that rely solely on textual information face inherent limitations, primarily due to the modality gap that exists within VLMs.
To bridge this gap, a growing body of research \cite{shin2022reco, shipard2023diversity, wallingford2023neural} introduces visual signals by either retrieving relevant images from external datasets or synthesizing them using generative models, as shown in Fig.~\ref{fig:data-free-methods} (b).

Retrieval-based methods attempt to provide visual grounding by leveraging large-scale unlabeled datasets. For example, ReCo \cite{shin2022reco} retrieves semantically relevant images using CLIP \cite{radford2021learning} from an external corpus and computes a reference image embedding for each category.
This embedding is then used to guide the recognition of corresponding image patches and refine the original zero-shot dense predictions.
Neural Priming \cite{wallingford2023neural} introduces a new classification head by computing the centroids of retrieved image sets for each category from the pre-training dataset, subsequently integrating this head with the original zero-shot classifier to enhance recognition.
Generative approaches extend this direction by synthesizing visual data to simulate examples for downstream tasks.
For instance, Shipard et al.\cite{shipard2023diversity} construct a synthetic training set using diffusion models \cite{rombach2022high}, generating a diverse set of images that provide rich visual cues in the absence of real data.
AttrSyn \cite{wang2025attributed} further boosts image diversity by leveraging LLMs to generate a wide range of attributes, which guide the generative model to produce class-consistent and discriminative samples.
SuS-X \cite{udandarao2023sus} combines both retrieval and generation strategies by introducing a visual support set, either constructed from large-scale datasets or generated via advanced diffusion models \cite{rombach2022high}. This support set enables information integration and provides auxiliary supervision during inference.
Finally, in the context of continual learning, GenCL \cite{seo2024just} generates synthetic images for novel classes using prompt-guided diffusion models, and then introduces an ensemble-based selector to curate a representative coreset from the generated samples, supporting robust and effective category representation over time.

\subsection{Network Modification}
Several data-free methods \cite{zhou2022extract, guo2023calip, lan2024clearclip, hajimiri2025pay} focus on modifying the network architecture of VLMs to enhance their suitability for downstream tasks, particularly dense prediction such as segmentation, as shown in Fig.~\ref{fig:data-free-methods} (c).
While these approaches are primarily developed for classification-oriented VLMs, their architectural enhancements significantly improve dense prediction performance.

A pioneering work, MaskCLIP \cite{zhou2022extract}, demonstrates that value embeddings in the final attention layers capture richer local information than global features, making them particularly effective for segmentation tasks. To further refine dense predictions from value embeddings, MaskCLIP introduces a key-based smoothing strategy and a denoising technique.
Building on this, CALIP \cite{guo2023calip} facilitates interaction between visual and textual features through a parameter-free attention module, and achieves improved classification results by ensembling outputs from multiple feature representations.
CLIP Surgery \cite{li2025closer} advances segmentation by introducing value-value attention to enhance local feature consistency and employing a feature surgery strategy to suppress noisy activations, thus improving both segmentation accuracy and interpretability.
GEM \cite{bousselham2024grounding} generalizes value-value attention to an any-any attention mechanism, enhancing consistency across groups of similar tokens by ensembling outputs from the modified attention applied to key, query, and value embeddings at every transformer layer.
SCLIP \cite{wang2024sclip} introduces a correlative self-attention mechanism, which yields spatially covariant features that better preserve fine local details.
Finally, ProxyCLIP \cite{lan2024proxyclip} presents a training-free framework that enhances CLIP’s open-vocabulary segmentation by integrating spatially consistent proxy attention maps generated from vision foundation models such as DINO \cite{caron2021emerging} and SAM \cite{kirillov2023segment}.

\section{Unsupervised Domain Transfer}
\label{sec:unsup}

\begin{figure*}[tbp]
\centering
\begin{minipage}[t]{0.32\textwidth}
  \centering
  \includegraphics[width=\linewidth]{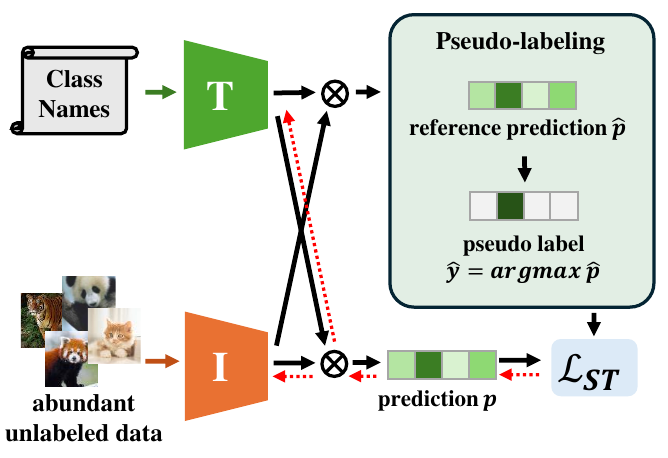}
  \vspace{-0.5cm}
  \caption*{(a) Self-training.}
\end{minipage}
\hfill
\begin{minipage}[t]{0.32\textwidth}
  \centering
  \includegraphics[width=\linewidth]{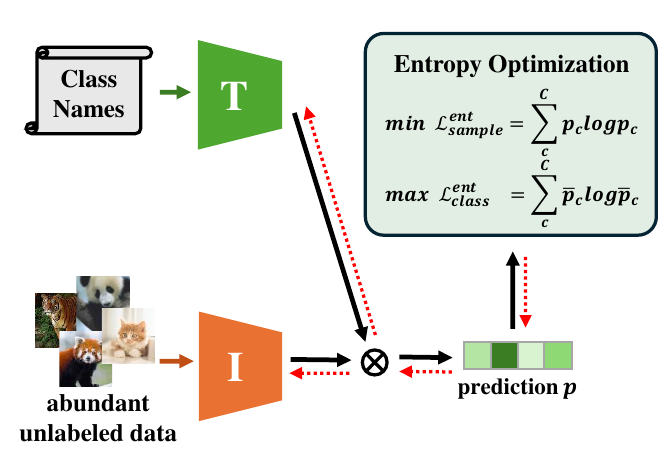}
  \vspace{-0.5cm}
  \caption*{(b) Entropy optimization.}
\end{minipage}
\hfill
\begin{minipage}[t]{0.32\textwidth}
  \centering
  \includegraphics[width=\linewidth]{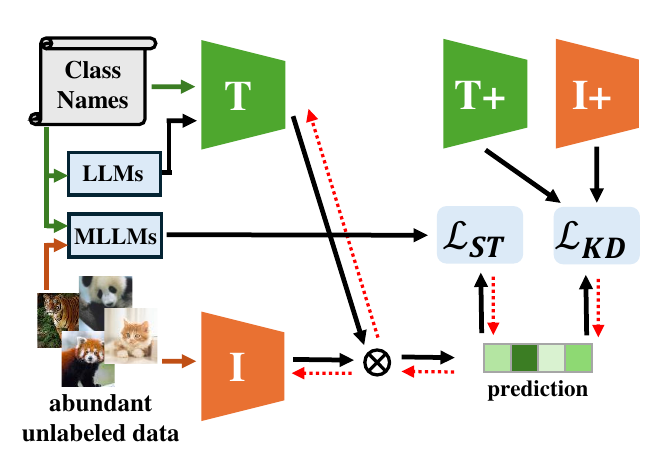}
  \vspace{-0.5cm}
  \caption*{(c) External resource utilization.}
\end{minipage}
 \vspace{-0.1cm}
\caption{Three representative strategies of the \textbf{unsupervised domain transfer} paradigm.}
\label{fig:unsupervised-methods}
\end{figure*}

\begin{table}[tbp]
\centering
\caption{Popular strategies along with their representative works of \textbf{unsupervised domain transfer}.}
\label{tab:unsuperviseddomain}
\resizebox{\linewidth}{!}{
\begin{tabular}{ll}
\toprule
\textbf{Strategies} & \textbf{Representative Works} \\
\midrule
\multirow{2}{*}{\textbf{Self-Training}} & UPL~\cite{huang2022unsupervised}, LaFTer~\cite{mirza2023lafter}, \\
& MUST~\cite{li2023masked}, ReCLIP~\cite{hu2024reclip}. \\
\midrule
\multirow{2}{*}{\textbf{Entropy Optimization}} & POUF~\cite{tanwisuth2023pouf}, CDBN~\cite{li2024data}, \\
& UEO~\cite{liang2024realistic}, TFUP-T~\cite{long2024training}. \\
\midrule
\multirow{2}{*}{\textbf{External Resource Utilization}} & Neural Priming~\cite{wallingford2023neural}, PEST~\cite{huang2023prompt}, \\
& PromptKD~\cite{li2024promptkd}, OTFusion~\cite{xu2025otfusion}. \\
\midrule
\multirow{2}{*}{\textbf{Miscellaneous}} & FARE~\cite{schlarmann2024robust},
OTTER~\cite{shin2024otter} \\
& ZLaP~\cite{kalantidis2024label}, TransCLIP~\cite{zanella2024boosting}. \\
\bottomrule
\end{tabular}
}
\label{tab:unsupervised-methods}
\end{table}

\textbf{Paradigm description.}
Unsupervised domain transfer for VLMs refers to the adaptation of pre-trained models to downstream tasks with \emph{abundant unlabeled data}.
Compared with data-free transfer, unsupervised domain transfer can use the unlabeled data of downstream tasks to better grasp the data distribution and thus achieve better performance.
The challenges of this paradigm mainly come from the filtering and processing of unlabeled data and the unsupervised alignment of VLM and unlabeled data.

We review existing unsupervised domain transfer methods and categorize their strategies into three primary approaches: self-training, entropy optimization, and external resource utilization. These categories are summarized in Table~\ref{tab:unsupervised-methods}, and we introduce each strategy in detail along with related methods in the following subsections.

\subsection{Self-Training}

Self-training is a widely used strategy in unsupervised learning, as shown in Fig.~\ref{fig:unsupervised-methods} (a), where the ground truth of the training data is absent.
Using this approach, unsupervised algorithms always manage to calculate high-quality pseudo labels on unlabeled samples as supervision signals.
How to obtain and iteratively refine pseudo labels to make VLM better adapts to the distribution of the unlabeled data is the key challenge for those methods.

UPL \cite{huang2022unsupervised} is one of the earliest efforts to explore unsupervised domain transfer for VLMs.
It selects a small set of high-confidence unlabeled samples for each category and optimizes the prompt parameters using a pseudo-labeling strategy,
\begin{equation}
    \label{eq:p-tuning2}
    \text{arg} \min_{\mathbf{p}}\mathbb{E}_{(\mathbf{x}, \hat{y})\sim\mathcal{D_{\text{select}}}} \mathcal{L}_{CE}(p({\mathbf{x}}), \hat{y
    }),
\end{equation}
where $D_{\text{select}}$ represents the selected high-confidence samples and $\mathcal{L}_{CE}$ denotes the cross-entropy loss.
This selective pseudo-labeling approach is later adopted by a number of subsequent works \cite{ma2024swapprompt, zhang2023unsupervised, li2024data, imam2024clip}.
SwapPrompt \cite{ma2024swapprompt} extends this self-training strategy with another swapped prediction mechanism, letting the two augmented views of the same image provide soft pseudo-label optimization supervision through an EMA-updated prompt for each other.
RS-CLIP \cite{li2023rs} introduces a curriculum learning framework, beginning with a small subset of high-confidence samples for self-training and progressively incorporating more data as optimization progresses, thus mitigating the noise from early-stage pseudo labels.
GTA-CLIP \cite{saha2025generate} proposes a transductive inference approach to pseudo-labeling, using iteratively refined, attribute-augmented similarities between image and text embeddings to improve label quality.
In a related approach, CPL \cite{zhang2024candidate} refines candidate pseudo labels by introducing both intra- and inter-instance label to reduce the negative impact of incorrect hard pseudo labels typically produced by VLMs.
Another work \cite{menghini2023enhancing} systematically investigates pseudo-labeling strategies across several unsupervised settings and demonstrates the effectiveness of pseudo-labeling in promoting more balanced and robust performance across diverse categories.

Inspired by FixMatch \cite{sohn2020fixmatch}, several unsupervised domain transfer methods \cite{mirza2023lafter, li2024data} apply both weak and strong augmentations to unlabeled data to enhance consistency learning in VLMs. These approaches typically generate pseudo labels using the weakly augmented views and consider them as supervisory signals for self-training on the strongly augmented ones.
LaFTer \cite{mirza2023lafter} leverages large language models (LLMs) to generate diverse textual data for training a text classifier, which in turn produces high-quality pseudo labels with weak augmented views for effective self-training.
MedUnA \cite{rahman2025can} proposes a dual-branch architecture, consisting of weak and strong branches for the visual encoder, and jointly optimizes them using a pseudo-labeling objective to enhance medical image classification.
NoLA \cite{imam2024clip} employs a DINO-based labeling network fed with weak augmentations to improve pseudo-label quality for training visual prompts.
DPA \cite{ali2025dpa} introduces dual prototype representations for both visual and textual branches, integrating their outputs to generate more robust pseudo labels.
Additionally, LP-CLIP \cite{laroudie2023improving} incorporates confidence estimates into the pseudo-labeling objective, thereby improving both classification accuracy and calibration.

Rather than filtering high-confidence samples and enhancing consistency between different augmentations, there are also some methods that generate pseudo labels in other ways.
MUST \cite{li2023masked} maintains an EMA model to produce high-qulity pseudo labels and incorporates a masked image modeling strategy to improve local image representation learning.
PEST \cite{huang2023prompt} enhances pseudo-label quality by ensembling predictions from multiple textual prompts and visually augmented views.
ReCLIP \cite{hu2024reclip} learns a projection space to better align visual and textual features and employs self-training with pseudo labels refined using Label Propagation \cite{iscen2019label}.
NtUA \cite{ali2023noise} constructs a confidence-weighted key-value cache of pseudo-labeled features and refines it through knowledge distillation, effectively mitigating label noise in scenarios with limited unlabeled data.
Similarly, TFUP-T \cite{long2024training} improves pseudo-label quality by building a cache model with representative samples and refining predictions based on both feature-level and semantic-level similarities.
To address the issue of low-confidence pseudo labels, FST-CBDG \cite{yan2024lightweight} employs soft pseudo labels and updates them using a moving average strategy during self-training.
For regression tasks, CLIPPR \cite{kahana2022improving} trains an adapter for the image encoder using zero-shot pseudo labels, optimizing performance by minimizing the distance between predicted and prior label distributions.

\subsection{Entropy Optimization}
Entropy optimization is a classic unsupervised learning objective, encouraging the model to make confident predictions on unlabeled data, as shown in Fig.~\ref{fig:unsupervised-methods} (b).
Unlike self-training, entropy optimization is not affected by erroneous pseudo-labels and behaves more stably on low-performance tasks.
Many algorithms minimize sample-level entropy, adapting the model to the unlabeled data distribution \cite{tanwisuth2023pouf, li2024data}, and also maximize category-level marginal entropy to avoid mode collapse \cite{tanwisuth2023pouf, long2024training, li2024data}.

POUF \cite{tanwisuth2023pouf} and CDBN \cite{li2024data} optimize textual prompt parameters with sample-level entropy minimization and category-level marginal entropy maximization.
An optimal transport objective is also incorporated into POUF for better alignment between the distribution of textual prototypes and the unlabeled data.
In order to improve both generalization and out-of-distribution detection performance of VLMs, UEO \cite{liang2024realistic} proposes universal entropy, utilizing marginal prediction instead of sample prediction for entropy maximization to stabilize the optimization process.

\subsection{External Resource Utilization}
Several recent approaches enhance the performance of VLMs by incorporating external resources beyond the available unlabeled data.
These resources often include retrieval-based image augmentation, introduction of (multimodal) large language models (MLLMs), and knowledge distillation from powerful VLMs or vision models, as shown in Fig.~\ref{fig:unsupervised-methods} (c).

Neural Priming \cite{wallingford2023neural} adopts a transductive learning paradigm by constructing a retrieval set of images based on category names. For each unlabeled sample, it selects the most visually similar images to form a fine-tuning dataset, thereby adapting the VLM to the target domain.
LaFTer \cite{mirza2023lafter} leverages GPT-3 \cite{brown2020language} to generate diverse textual descriptions, which are then used to train a text classifier tailored for the downstream task.
Similarly, PEST \cite{huang2023prompt} and GTA-CLIP \cite{saha2025generate} query LLMs such as GPT-3 \cite{brown2020language} and LLaMA \cite{touvron2023llama} to create multiple prompts per class. These prompts improve pseudo-label quality through ensemble-based prompt inference.
LatteCLIP \cite{cao2025latteclip} utilizes LLaVA \cite{liu2023visual} to generate image captions, which support more accurate textual prototype construction for VLM adaptation.

In the context of knowledge distillation, PromptKD \cite{li2024promptkd} reuses textual features from a larger teacher VLM to guide the training of an image encoder, thereby transferring semantic knowledge.
Going a step further, KDPL \cite{mistretta2024improving} jointly optimizes prompts in both visual and textual input spaces, balancing performance and efficiency.
NtUA \cite{ali2023noise} improves pseudo-label reliability by incorporating the image encoder of a stronger VLM, enhancing both label quality and confidence estimation.
OTFusion~\cite{xu2025otfusion} aligns VLMs' embedding with features extracted from powerful vision models (e.g., DINO) via optimal transport to obtain refined predictions.

\subsection{Miscellaneous}
There are also several recent approaches that address unsupervised domain transfer for VLMs through diverse strategies \cite{schlarmann2024robust, benigmim2025floss}.
ZPE~\cite{allingham2023simple} designs a prompt ensembling strategy that leverages unlabeled data to address frequency biases in words and concepts and assigns appropriate ensembling weights to multiple prompt templates.
uCAP \cite{nguyen2024ucap} formulates image generation as a function of class names and latent, domain-specific prompts. It employs an energy-based likelihood framework to infer optimal prompts from unlabeled data.
To enhance adversarial robustness while preserving performance on clean inputs, FARE \cite{schlarmann2024robust} optimizes the vision encoder to align the features of adversarially perturbed images with those of their clean counterparts as computed by the original VLM.
OTTER \cite{shin2024otter} addresses label distribution mismatch by leveraging optimal transport to align model predictions with an estimated label distribution in the target domain.
Moreover, InMaP \cite{qian2023intra} learns class proxies directly in the vision space using refined pseudo labels derived from text embeddings, thereby narrowing the modality gap between visual and textual representations in VLMs.
Subsequent methods exploit various strategies for modeling vision-text features, including label propagation \cite{kalantidis2024label} and Dirichlet distributions \cite{martin2024transductive}.
TransCLIP \cite{zanella2024boosting} proposes a plug-and-play transductive framework that optimizes a KL-regularized objective with an efficient block Majorize-Minimize algorithm, integrating the text-encoder knowledge together.

\section{Episodic Test-Time Adaptation}
\label{sec:insta}

\begin{figure*}[tbp]
\centering
\begin{minipage}[t]{0.33\textwidth}
  \centering
  \includegraphics[width=\linewidth]{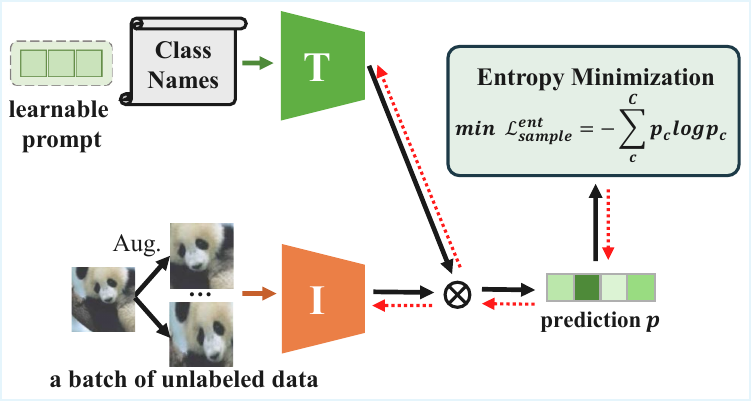}
  \vspace{-0.5cm}
  \caption*{(a) Entropy minimization.}
\end{minipage}
\hfill
\begin{minipage}[t]{0.29\textwidth}
  \centering
  \includegraphics[width=\linewidth]{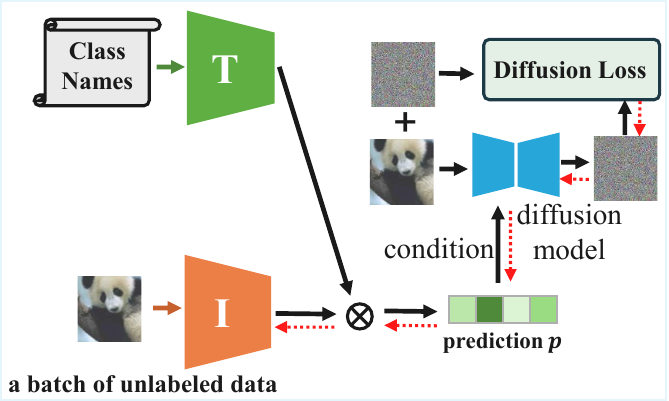}
  \vspace{-0.5cm}
  \caption*{(b) Feedback signal.}
\end{minipage}
\hfill
\begin{minipage}[t]{0.32\textwidth}
  \centering
  \includegraphics[width=\linewidth]{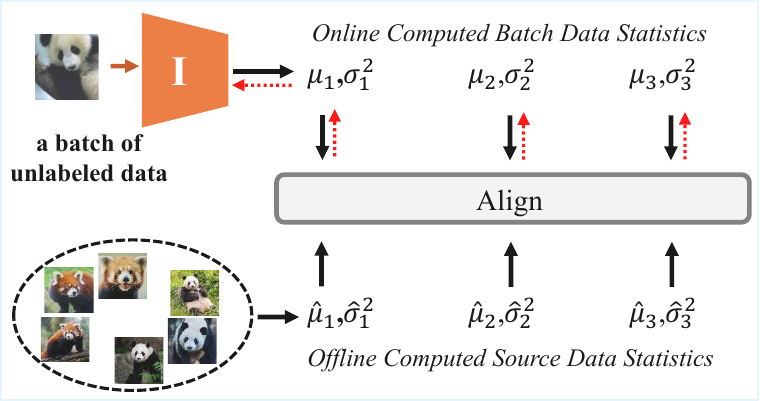}
  \vspace{-0.5cm}
  \caption*{(c)  Distribution alignment.}
\end{minipage}
\vspace{-0.1cm}
\caption{Three representative strategies of the \textbf{episodic test-time adaptation} paradigm.}
\label{fig:tpt}
\end{figure*}

\begin{table}[tbp]
\centering
\caption{Popular strategies along with their representative works of \textbf{episodic test-time adaptation}.}
\renewcommand{\arraystretch}{1.2}
\begin{tabular}{@{} p{3.2cm} p{4.4cm} @{}}
\toprule
\textbf{Strategies} & \textbf{Representative Works} \\
\midrule

 \multirow{2}{*}{\makecell[l]{\textbf{Entropy Minimization}}} & TPT~\cite{shu2022test}, DiffTPT~\cite{feng2023diverse}, \\
   & R-TPT~\cite{sheng2025r}, DTS-TPT~\cite{yan2024dts}. \\
  \hline
  \multirow{2}{*}{\makecell[l]{\textbf{Feedback Signal}}} & Diffusion-TTA~\cite{prabhudesai2023diffusion}, \\
   & RLCF~\cite{zhao2023test}, BPRE~\cite{qiao2025bidirectional} . \\
  \hline
   \multirow{2}{*}{\makecell[l]{\textbf{Distribution Alignment}}} & PromptAlign~\cite{abdul2023align}, MTA~\cite{zanella2024test}, \\
  & TAPT~\cite{wang2024tapt}, StatA~\cite{zanella2025realistic}. \\
  \hline
   \multirow{2}{*}{\makecell[l]{\textbf{Self-Supervised Learning}}} & Self-TPT~\cite{zhu2024efficient}, InCPL~\cite{yin2024context}, \\
  & LoRA-TTT~\cite{kojima2025lora}, T3AL~\cite{liberatori2024test}. \\
  \hline
   \multirow{2}{*}{\makecell[l]{\textbf{Miscellaneous} }} & AWT~\cite{zhu2024awt}, RA-TTA~\cite{lee2025ratta}, \\
  & SCAP~\cite{zhang2025scap}, ZERO~\cite{farina2024frustratingly}. \\
\bottomrule
\end{tabular}
\label{tab:ins}
\end{table}

\textbf{Paradigm description.}
Episodic test-time adaptation is a popular learning paradigm in which a pre-trained VLM is \emph{adapted at inference time using a single batch of unlabeled test data}. The goal is to leverage the knowledge embedded in the pre-trained VLM to accurately predict labels for the current batch, without requiring access to multiple test batches or labeled data during adaptation.

We review existing episodic test-time adaptation methods and categorize their strategies into four primary approaches: entropy minimization, feedback signal, distribution alignment, and self-supervised learning. These categories are summarized in Table~\ref{tab:ins}, and we introduce each strategy in detail along with related methods in the following subsections.

\subsection{Entropy Minimization}

Entropy minimization is a widely adopted strategy for TTA~\cite{liang2024comprehensive} by adjusting the model's parameters to make its output predictions more confident with lower entropy, as shown in \cref{fig:tpt} (a). This process encourages the model to produce low-uncertainty outputs for the test data, often improving its performance under distribution shifts.

Shu et al.~\cite{shu2022test} introduced test-time prompt tuning (TPT) as the first method for adapting pre-trained VLMs at test time. TPT optimizes a text prompt $\mathbf{p} = [V]_1[V]_2...[V]_M$ for each test sample using entropy minimization, combined with confidence selection to ensure consistent predictions across augmented views. Specifically, TPT generates $N$ randomly augmented views of a test image $\mathbf{x}$ using a set of random augmentations $\mathcal{A}$ and minimizes the entropy of the averaged prediction probability distribution: 
\begin{align}
\label{eq:ent-min}
    &\mathbf{p}^{\ast} = \text{arg}\min_{\mathbf{p}}-\sum_{i=1}^{C}\tilde{p}_{\mathbf p} (y_i|\mathbf{x}) \log \tilde{p}_{\mathbf p}(y_i|\mathbf{x}), \\
    &\tilde{p}_{\mathbf p}(y_i|\mathbf{x}) = \frac{1}{\rho N}\sum_{i=1}^{N} \mathbb{I}[\mathbf{H}(p_i) \le \eta] p_{\mathbf{p}}(y|\mathcal{A}_i(\mathbf{x})).
\end{align}
Here, $p_{\mathbf{p}}(y|\mathcal{A}_i(\mathbf{x}))$ represents the class probability vector for the $i$-th augmented view of $\mathbf{x}$ under prompt $\mathbf{p}$. TPT selects $\rho$-percentile confident samples with a prediction entropy below a threshold $\eta$ to filter out noisy predictions, using a confidence mask $\mathbb{I}[\mathbf{H}(p_i) \le \eta]$, where $\mathbf{H}$ denotes the entropy of predictions on augmented samples. 
Instead of applying random augmentations as in TPT, DiffTPT~\cite{feng2023diverse,feng2025diffusion} leverages pre-trained diffusion models to generate diverse augmentations and employs cosine similarity-based filtering to remove spurious samples. 
R-TPT~\cite{sheng2025r} employs a reliability-based weighted ensembling strategy to aggregate information from trustworthy augmented views of the test sample.
C-TPT~\cite{yoon2024c} optimizes prompts by maximizing text feature dispersion, observing that better-calibrated predictions correlate with higher text feature dispersion. O-TPT~\cite{sharifdeen2025tpt} improves calibration by enforcing orthogonality constraints on class-specific textual prompt features during tuning to maximize their angular separation. 
Furthermore, DTS-TPT~\cite{yan2024dts} extends TPT to video data for zero-shot activity recognition.

Beyond tuning text prompts to minimize entropy, several works also explore visual prompts~\cite{sun2023vpa}, multimodal prompts~\cite{abdul2023align}, low-rank attention weights~\cite{imam2024test}, and learnable noise~\cite{imam2025noise}. 
For example, PromptAlign~\cite{abdul2023align} uses multimodal prompt learning to align image token distributions between a pre-computed source proxy dataset and test samples. 
TTL~\cite{imam2024test} adapts low-rank attention weights during test time through a confidence maximization objective, enabling efficient adaptation without altering prompts or backbone parameters.

\subsection{Feedback Signal}
Some studies explore leveraging feedback signals from diffusion~\cite{prabhudesai2023diffusion} or CLIP-like models~\cite{zhao2023test,qiao2025bidirectional} for TTA, as shown in \cref{fig:tpt} (b).
For example, Diffusion-TTA~\cite{prabhudesai2023diffusion} leverages generative feedback from diffusion models to adapt pre-trained discriminative models at test time by optimizing image likelihood, significantly improving performance across tasks like classification, segmentation, and depth prediction. Diffusion-TTA consists of discriminative and generative modules. Given an image $\mathbf{x}$, the discriminative model $f_\theta$ predicts task output $y$ (\cref{eqn:clip-metric} for VLMs).  The task output $y$ is transformed into condition $\mathbf{c}$. 
For image classification, $y$ represents a probability distribution over $C$ categories, $y \in [0,1]^C, y^\top \mathbf{1}_C=1$. Given the learned text embeddings of a text-conditional diffusion model for the $C$  categories $ \mathbf{t}_j \in \mathbb{R}^d, j \in \{1...C\}$, the diffusion condition is $\mathbf{c} = \sum_{j=1}^C y_j \cdot \mathbf{t}_j$.   
Finally, the generative diffusion model $\epsilon_\phi$ is used to measure the likelihood of the input image, conditioned on $\mathbf{c}$. This consists of using the diffusion model $\epsilon_\phi$ to predict the added noise $\epsilon$ from the noisy image $\mathbf{x}_t$ and condition $\mathbf{c}$. The image likelihood is maximized using diffusion loss by updating the discriminative and generative model weights via backpropagation:
\begin{equation}\label{e:classifier_adapt}
\mathcal{L}_{\text{diff}} = \mathbb{E}_{t, \epsilon}
\| \epsilon_{\phi}(\sqrt{\bar{\alpha}_t} x + \sqrt{1-\bar{\alpha}_t} \epsilon, \mathbf{c} , t) - \epsilon \|^2,
\end{equation}
where $\bar{\alpha}_t$ defines how much noise is added at each time step~$t$.
Differently, RLCF~\cite{zhao2023test} utilizes CLIP-based feedback through reinforcement learning and employs CLIPScore~\cite{hessel2021clipscore} as a reward signal to provide feedback for VLMs.
BPRE~\cite{qiao2025bidirectional} mitigates text-conditioned bias by using a quality-aware reward module based on intrinsic visual features, forming a self-evolving feedback loop with prototype refinement to enhance adaptation to distribution shifts.

\subsection{Distribution Alignment}
Distribution alignment methods align test sample distributions with known source characteristics or refine representations for improved consistency~\cite{sui2024just}, as shown in \cref{fig:tpt} (c).
For example, PromptAlign~\cite{abdul2023align} bridges the source-to-target distribution gap by jointly updating multimodal prompts to align the per-layer image-token statistics of augmented test views with offline-computed source statistics through a combined alignment and entropy minimization loss.
To enhance adversarial robustness, TAPT~\cite{wang2024tapt} adapts statistical alignment by using a loss function that, at inference, aligns the augmented visual embeddings of a test sample with pre-computed statistics from both clean and adversarially perturbed images.
StatA~\cite{zanella2025realistic} preserves text encoder knowledge during adaptation by utilizing statistical anchors that penalize deviations from text-derived Gaussian priors.

Complementary to these approaches, MTA~\cite{zanella2024test} employs a robust MeanShift algorithm to identify density modes in the feature space while concurrently optimizing them with an inlierness score to automatically assess each view's quality. 
Beyond global distribution alignment, several methods focus on class-aware prototype alignment. PromptSync~\cite{khandelwal2024promptsync} performs class-aware prototype alignment of the test sample with source class prototypes, weighted by mean class probabilities derived from confident augmented views.
Also utilizing class prototypes, TPS~\cite{sui2024just} pre-computes class prototypes and then, for each test sample, dynamically learns shift vectors to adjust these prototypes directly within the shared embedding space.

\subsection{Self-Supervised Learning}

Self-supervised learning~\cite{gui2024survey,hendrycks2019using} is a powerful technique for learning transferable representations.
Self-TPT~\cite{zhu2024efficient} introduces contrastive prompt tuning as a self-supervised learning strategy, which aims to minimize intra-class distances while maximizing inter-class separation by leveraging contrastive learning principles. Specifically, for each class token, multiple prompt variations are generated by altering the insertion point of the class token (e.g., beginning, middle, or end of the prompt sequence). This creates positive pairs from the same class and negative pairs from different classes, encouraging the model to learn more robust class representations. The contrastive loss is formulated as:
\begin{equation}
\mathcal{L} = -\sum_{i=1}^{4C} \log \frac{\sum_{j \in P(i)} \exp \left( \frac{\mathbf{t}_i \cdot \mathbf{t}_j}{\tau} \right)}{\sum_{j=1, j \neq i}^{4C} \exp \left( \frac{\mathbf{t}_i \cdot \mathbf{t}_j}{\tau} \right)},
\end{equation}
where \( \mathbf{t}_i \) and \( \mathbf{t}_j \) are the projected text features of different views, \( P(i) \) denotes the set of positive samples for view \( i \), and \( \tau \) is a temperature parameter.
In contrast, LoRA-TTT~\cite{kojima2025lora} updates only the low-rank parameters in the image encoder using a memory-efficient reconstruction loss, computed as the mean squared error of class tokens from top-confidence augmented and masked views, to enhance global feature understanding.
In addition, InCPL~\cite{yin2024context} enables efficient model adaptation by optimizing visual prompts from a few labeled examples through a context-aware unsupervised loss and a cyclic learning strategy.
T3AL~\cite{liberatori2024test} generates and refines temporal action proposals by first deriving video-level pseudo-labels from a pretrained VLM, then using a self-supervised method to create initial proposals, and finally enhancing them with frame-level textual descriptions.

\subsection{Miscellaneous}
Beyond the previously discussed approaches, additional techniques have been developed for episodic test-time adaptation of VLMs~\cite{liu2025test,cai2025local,yi2024leveraging,an2023perceptionclip,ge2023improving,shen2024diffclip,rahaman2025leveraging,li2024wavedn}. One line of work uses retrieval-based strategies~\cite{eom2023cross,lee2025ratta,conti2024vocabulary}. For instance, X-MoRe~\cite{eom2023cross} retrieves relevant captions via a two-step cross-modal retrieval and ensembles image and text predictions using dynamically weighted modal-confidence scores. RA-TTA~\cite{lee2025ratta} utilizes fine-grained text descriptions to guide a two-step retrieval of relevant external images, which are then used in a description-based adaptation process to refine the model's initial prediction. 
Another line of work tries to improve the calibration of VLMs~\cite{murugesan2024robust,farina2024frustratingly,yoon2024c,sharifdeen2025tpt}. Besides retrieval and calibration, other representative work includes optimal transport~\cite{zhu2024awt}, spurious features erasing~\cite{ma2024invariant}, loss landscape~\cite{li2025test}, counterattack~\cite{xing2025clip}, and supportive cliques~\cite{zhang2025scap}.
Several methods focus on improving CLIP’s dense prediction abilities for open-vocabulary semantic segmentation by addressing its image-level pre-training limitations~\cite{wysoczanska2024clip,shao2024explore,kang2024defense,shin2022reco,sun2024training,hajimiri2025pay,bousselham2024grounding}.
The external knowledge, such as MLLMs and LLMs, can also be used during inference, without requiring additional training or fine-tuning on task-specific data~\cite{li2024visual,yin2024s,abdelhamed2024you,wei2024enhancing,munir2025tlac,miller2025sparc,esfandiarpoor2023follow,guo2022texts}.

\section{Online Test-Time Adaptation}
\label{sec:online}

\begin{figure*}[tbp]
\centering
\begin{minipage}[t]{0.3\textwidth}
  \centering
  \includegraphics[width=\linewidth]{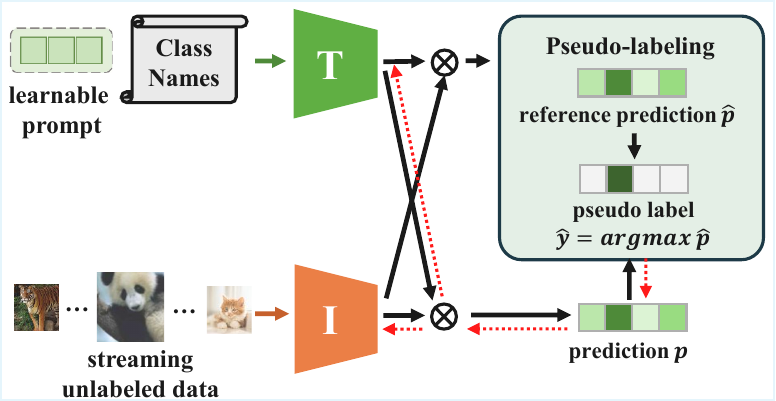}
  \vspace{-0.5cm}
  \caption*{(a) Pseudo-labeling.}
\end{minipage}
\hfill
\begin{minipage}[t]{0.335\textwidth}
  \centering
  \includegraphics[width=\linewidth]{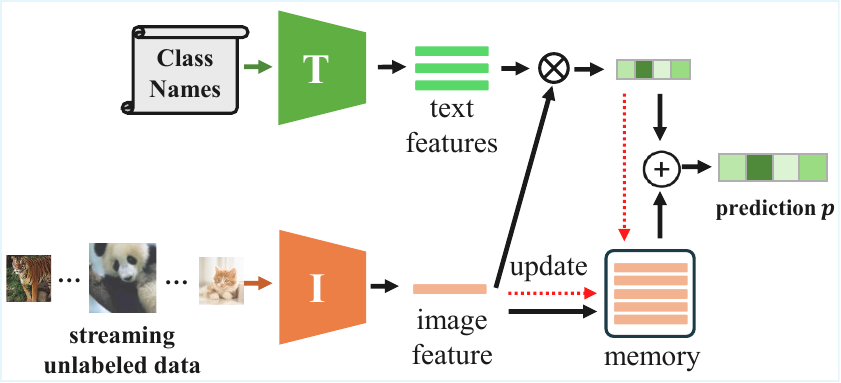} \vspace{-0.5cm}
  \caption*{(b) Memory mechanisms.}
\end{minipage}
\hfill
\begin{minipage}[t]{0.28\textwidth}
  \centering
  \includegraphics[width=\linewidth]{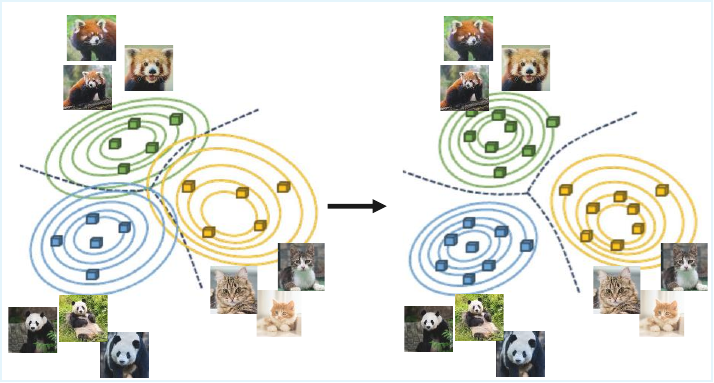}
  \vspace{-0.5cm}
  \caption*{(c)  Distribution modeling.}
\end{minipage}
\vspace{-0.1cm}
\caption{Three representative strategies of the \textbf{online test-time adaptation} paradigm.}
\label{fig:online}
\end{figure*}

\textbf{Paradigm  description.} 
Online test-time adaptation is another TTA paradigm designed for \emph{streaming data scenarios, where unlabeled data arrives sequentially in mini-batches}. Given a pre-trained VLM, the objective is to adapt the model online to each incoming mini-batch in order to accurately predict its labels under potential distribution shifts. Unlike episodic adaptation, which adapts to each batch independently, online adaptation continuously updates the model by leveraging knowledge accumulated from previously observed mini-batches. This enables more effective and efficient label prediction in dynamic, streaming environments.

We review existing online test-time adaptation methods and categorize their strategies into three primary approaches: pseudo-labeling, memory mechanisms, and distribution modeling. These categories are summarized in Table~\ref{tab:online}, and we introduce each strategy in detail along with related methods in the following subsections.

\begin{table}[tbp]
\centering
\caption{Popular strategies along with their representative works of \textbf{online test-time adaptation}.}
\renewcommand{\arraystretch}{1.2}
\begin{tabular}{@{}  p{3.2cm} p{3.9cm} @{}}
\toprule
\textbf{Strategies} & \textbf{Representative Works} \\
\midrule
\multirow{2}{*}{\makecell[l]{\textbf{Pseudo-Labeling}}} & DART~\cite{liu2024dart}, CLIPArTT~\cite{hakim2024clipartt}, \\
& CLIP-OT~\cite{mishra2024words}, WATT~\cite{osowiechi2024watt}. \\
\hline
\multirow{2}{*}{\makecell[l]{\textbf{Memory Mechanisms}}} & TDA~\cite{karmanov2024efficient}, DMN~\cite{zhang2024dual2}, \\
& DPE~\cite{zhang2024dual}, BaFTA~\cite{hu2024bafta}. \\
\hline
\multirow{2}{*}{\makecell[l]{\textbf{Distribution Modeling} }} & OGA~\cite{fuchs2025online}, DOTA~\cite{han2024dota}, \\
& BCA~\cite{zhou2025bayesian}, DN~\cite{zhou2024test}. \\
\hline
\multirow{2}{*}{\makecell[l]{\textbf{Miscellaneous }}} & DynaPrompt~\cite{xiao2025dynaprompt}, TCA~\cite{wang2024less}, \\
& ECALP~\cite{li2025efficient}, OnZeta~\cite{qian2024online}. \\
\bottomrule
\end{tabular}
\label{tab:online}
\end{table}

\subsection{Pseudo-Labeling}
Pseudo-labeling assigns class labels to unlabeled test samples and optimizes the cross-entropy loss between predictions and pseudo-labels to guide model adaptation, as shown in \cref{fig:online} (a). However, due to distribution shifts, pseudo-labels may be noisy, which can negatively impact learning. Various methods have been proposed to mitigate this issue. Many methods refine the pseudo-labeling process itself; for instance, IST~\cite{ma2024improved} employs graph-based correction and non-maximum suppression for pseudo-label refinement, stabilizing updates with parameter moving averages. Others, like CLIPArTT~\cite{hakim2024clipartt} dynamically construct text prompts from top-\( K \) predicted classes to serve as pseudo-labels, while 
CLIP-OT~\cite{mishra2024words} utilizes optimal transport for label assignment alongside multi-template knowledge distillation. CTPT~\cite{wang2024ctpt} focuses on iterative prompt updates guided by stable class prototypes and accurate pseudo-labels. SwapPrompt~\cite{ma2024swapprompt} proposes a dual-prompt and swapped prediction mechanism for efficient prompt adaptation. 

Several approaches enhance pseudo-labeling by integrating it with other mechanisms. For instance, SCP~\cite{wang2024towards} uses self-text distillation with conjugate pseudo-labels to improve robustness and minimize overfitting.  
WATT~\cite{osowiechi2024watt} combines diverse text templates, pseudo-label-based updates with periodic weight averaging, and text ensembling. 
To handle noisy target data, AdaND~\cite{cao2025noisy} introduces an adaptive noise detector trained with pseudo-labels from a frozen model to decouple noise detection from classification.
DART~\cite{liu2024dart} learns adaptive multimodal prompts (class-specific text and instance-level image) while retaining knowledge from prior test samples. 
ROSITA~\cite{sreenivas2024effectiveness} employs a contrastive learning objective with dynamically updated feature banks to enhance the discriminability of OOD samples.
Finally, TIPPLE~\cite{lu2025task} adopts a two-stage approach, first using online pseudo-labeling with an auxiliary text classification task and diversity regularization for task-oriented prompt learning, then refining this task-level prompt with a tunable residual for each test instance.

\subsection{Memory Mechanisms}




Memory-based methods leverage dynamic or static memory structures to store and retrieve feature representations and pseudo-labels from test samples, as shown in \cref{fig:online} (b). These methods enable progressive refinement of predictions by utilizing confident outputs and historical information, enhancing robustness and adaptability without requiring extensive retraining or backpropagation~\cite{huang2025cosmic,ding2025space,wang2024prompt,tong2024test}.
Inspired by Tip-Adapter~\cite{zhang2022tip}, Karmanov et al.~\cite{karmanov2024efficient} propose a training-free dynamic adapter (TDA) without requiring backpropagation. The core of TDA is a dynamic key-value cache system that stores pseudo-labels and corresponding feature representations from test samples. This cache enables progressive refinement of predictions by leveraging confident test-time outputs, facilitating efficient adaptation. 
Similarly, DMN~\cite{zhang2024dual2} leverages static memory for training data knowledge and dynamic memory for online test feature preservation. 
BoostAdapter~\cite{zhangboostadapter} leverages a lightweight key-value memory to retrieve features from instance-agnostic historical samples and instance-aware boosting samples.
HisTPT~\cite{zhang2024historical} constructs three complementary knowledge banks—local, hard-sample, and global—to preserve useful information from previously seen test samples. 
AdaPrompt~\cite{zhang2024robust} introduces a confidence-aware buffer that stores and utilizes only class-balanced, high-confidence samples to ensure the prompt updates are robust and stable.

Other works utilize dynamically evolving class prototypes to capture accurate multimodal representations during inference. By continuously updating these prototypes from unlabeled test samples, these methods enhance model adaptability, robustness, and efficiency~\cite{zhai2025mitigating,yi2025mint}.
For example, DPE~\cite{zhang2024dual} simultaneously evolves two sets of prototypes—textual and visual—to progressively capture accurate multimodal representations for target classes during test time. 
BaFTA~\cite{hu2024bafta} uses backpropagation-free online clustering to estimate class centroids and robustly aggregate class embeddings with visual-text alignment, guided by entropy-based reliability for improved zero-shot performance.
BATCLIP~\cite{maharana2024enhancing} introduces a projection matching loss to improve alignment between visual class prototypes and text features, and a separability loss to increase the distance between these prototypes for more discriminative features.

\subsection{Distribution Modeling}

Distribution modeling methods model the distribution of visual or multimodal features, often using Gaussian estimations, to refine predictions during inference~\cite{xiao2024any,cui2025bayestta,dai2025free}, as shown in \cref{fig:online} (c). By leveraging probabilistic frameworks and incorporating zero-shot priors, these methods enhance adaptability and robustness without requiring extensive hyperparameter tuning or backpropagation.
For instance, OGA~\cite{fuchs2025online}  models the likelihood of visual features using multivariate Gaussian distributions and incorporates zero-shot priors within a {maximum a posteriori} estimation framework. 
Similarly, DOTA~\cite{han2024dota} estimates Gaussian class distributions to compute Bayes-based posterior probabilities for adaptation, achieves fast inference without gradient backpropagation, and incorporates a human-in-the-loop mechanism to handle uncertain samples and enhance test-time performance. 
BCA~\cite{zhou2025bayesian} continuously updates text-based class embeddings to align likelihoods with incoming image features and concurrently refines class priors using the resulting posterior probabilities. 
On the other hand, DN~\cite{zhou2024test} approximates negative sample information using the mean representation of test samples, enhancing alignment with the model's optimization objective without requiring retraining or fine-tuning.

\subsection{Miscellaneous}
Beyond the previously discussed approaches, additional techniques have been developed for online test-time adaptation of vision-language models~\cite{chen2024test,wang2024less,lafon2025cliptta,han2025negation}. For instance, DynaPrompt~\cite{xiao2025dynaprompt} mitigates error accumulation in online adaptation by dynamically selecting and updating prompts per test sample based on entropy and confidence scores, while maintaining an adaptive buffer to add informative prompts and discard inactive ones. ECALP~\cite{li2025efficient} performs inference without task-specific tuning by dynamically expanding a graph over text prompts, few-shot examples, and test samples, using context-aware feature re-weighting to exploit the test sample manifold without requiring additional unlabeled data. OnZeta~\cite{qian2024online} sequentially processes test images for immediate prediction without storage, using online label learning to model the target distribution and online proxy learning to bridge the image-text modality gap via class-specific vision proxies.
Besides, other representative work includes support set~\cite{yan2025test}, token condensation~\cite{wang2024less}, prompt distillation~\cite{zhang2025hierarchical}, and more~\cite{tong2025zero,chen2025small,adachi2025uniformity,sarkar2024active}.

\section{Applications} 
\label{sec:application}





\setlength{\tabcolsep}{2.0pt}
\begin{table*}[t]
    \centering
    \caption{Overview of datasets from various tasks used in VLM-based unsupervised learning methods. (DFT=Data-Free Transfer, UDF=Unsuperivsed Domain Transfer, ETTA=Episodic Test-Time Adaptation, OTTA=Online Test-Time Adaptation)
    }
    \label{tab:dataset}
    \resizebox{0.825\textwidth}{!}{
        \begin{tabular}{llcccccc} 
        \toprule
        \multirow{2}{*}{Dataset} & \multirow{2}{*}{Task} & \multirow{2}{*}{\# Classes} & \multirow{2}{*}{\# Test sample} & \multicolumn{4}{c}{Popularity} \\
        & & & & DFT & UDF & ETTA & OTTA \\
        \midrule
        Caltech101~\cite{fei2004learning} & Object Classification & 100 & 2,465 & \starf & \starf & \starf & \starf \\
        OxfordPets~\cite{parkhi2012cats} & Object Classification \& OOD Detection & 37 & 3,669 & \starf & \starf & \starf & \starf \\
        StanfordCars~\cite{krause20133d} & Object Classification \& OOD Detection & 196 & 8,041 & \starf & \starf & \starf & \starf \\
        Flowers102~\cite{nilsback2008automated} & Object Classification & 102 & 2,463 & \starf & \starf & \starf & \starf \\
        Food101~\cite{bossard2014food} & Object Classification \& OOD Detection & 101 & 30,300 & \starf & \starf & \starf & \starf \\
        FGVCAircraft~\cite{maji2013fine} & Object Classification & 100 & 3,333 & \starf & \starf & \starf & \starf \\
        SUN397~\cite{xiao2010sun} & Object Classification \& OOD Detection & 397 & 19,850 & \starf & \starf & \starf & \starf \\
        DTD~\cite{cimpoi2014describing} & Object Classification \& OOD Detection & 47 & 1,692 & \starf & \starf & \starf & \starf \\
        EuroSAT~\cite{helber2019eurosat} & Object Classification & 10 & 8,100 & \starf & \starf & \starf & \starf \\
        UCF101~\cite{soomro2012ucf101} & Object Classification \& Action Recognition & 101 & 3,783 & \starf & \starf & \starf & \starf \\
        
        ImageNet~\cite{deng2009imagenet} & Object Classification \& OOD Detection & 1,000 & 50,000 & \starf & \starf & \starf & \starf \\
        ImageNet-A~\cite{hendrycks2021natural} & Object Classification & 200 & 7,500 & \starh & \starh & \starf & \starf \\
        ImageNet-V2~\cite{recht2019imagenet} & Object Classification & 1,000 & 10,000 & \starh & \starh & \starf & \starf \\
        ImageNet-R~\cite{hendrycks2021many} & Object Classification & 200 & 30,000 & \starh & \starh & \starf & \starf \\
        ImageNet-Sketch~\cite{wang2019learning} & Object Classification & 1,000 & 50,889 & \starh & \starh & \starf & \starf \\
        Office-Home~\cite{venkateswara2017deep} & Object Classification & 65 & 15,588 & \stare & \starh & \stare & \stare \\
        Office~\cite{saenko2010adapting} & Object Classification & 31 & 4,110 & \stare & \starh & \stare & \stare \\
        DomainNet~\cite{peng2019moment} & Object Classification & 345 & 176,743 & \stare & \starh & \stare & \stare \\

        PASCAL VOC 2012~\cite{everingham2015pascal} & Semantic Segmentation & 20 & 1,449 & \starf & \starf & \starf & \stare \\
        PASCAL Context~\cite{mottaghi2014role} & Semantic Segmentation & 59 & 5,105 & \starf & \starf & \starh & \stare \\
        COCO Stuff~\cite{caesar2018coco} & Semantic Segmentation & 172 & 4,172 & \starf & \starf & \starf & \stare \\
        ADE20K~\cite{zhou2019semantic} & Semantic Segmentation & 150 & 2,000 & \starf & \starh & \starh & \stare \\
        COCO-Object~\cite{lin2014microsoft} & Semantic Segmentation & 80 & 5,000 & \starf & \stare & \stare & \stare \\
        Cityscapes~\cite{cordts2016cityscapes} & Semantic Segmentation & 27 & 500 & \starf & \starh & \stare & \stare \\
        KITTI-STEP~\cite{weber2021step} & Semantic Segmentation & 19 & 2,981 & \stare & \stare & \stare & \stare \\
        FireNet~\cite{firenet} & Semantic Segmentation & - & 1,452 & \stare & \stare & \stare & \stare \\

        Bongard-HOI~\cite{jiang2022bongard} & Visual Reasoning & 2 & 13,914 & \stare & \stare & \starh & \stare \\

        CIFAR-100~\cite{krizhevsky2009learning} & OOD Detection & 100 & 10,000 & \starh & \stare & \stare & \stare \\
        CUB-200-2011~\cite{wah2011caltech} & OOD Detection & 200 & 5,794 & \starh & \stare & \stare & \stare \\
        iNaturalist~\cite{van2018inaturalist} & OOD Detection & 5,089 & 675,170 & \starh & \stare & \stare & \stare \\
        Places~\cite{zhou2017places} & OOD Detection & 365 & 18,250 & \starh & \stare & \stare & \stare \\
        ImageNet-O~\cite{hendrycks2021natural} & OOD Detection & 200 & 2000 & \starh & \stare & \stare & \stare \\
        OpenImage-O~\cite{krasin2017openimages} & OOD Detection & - & 17,632 & \starh & \stare & \stare & \stare \\
        MS-COCO~\cite{lin2014microsoft} & Text-Image Retrieval \& Image Captioning & - & 5,000 & \stare & \stare & \stare & \starh \\
        Flickr30K~\cite{plummer2015flickr30k} & Text-Image Retrieval \& Image Captioning & - & 1,000 & \stare & \stare & \stare & \starh \\
        Fashion-Gen~\cite{rostamzadeh2018fashion} & Text-Image Retrieval & - & 32,528 & \stare & \stare & \stare & \starh \\
        CUHK-PEDES~\cite{li2017person} & Text-Image Retrieval & - & 40,206 & \stare & \stare & \stare & \starh \\
        ICFG-PEDES~\cite{ding2021semantically} & Text-Image Retrieval & - & 54,522 & \stare & \stare & \stare & \starh \\
        Nocaps~\cite{agrawal2019nocaps} & Text-Image Retrieval \& Image Captioning & - & 15,100 & \stare & \stare & \stare & \starh \\
        
        Guangzhou Dataset~\cite{kermany2018identifying} & Medical Image Diagnosis & 2 & 5,856 & \starh & \starh & \stare & \stare \\
        Montgomery Dataset~\cite{jaeger2014two} & Medical Image Diagnosis & 2 & 138 & \starh & \starh & \stare & \stare \\
        Shenzhen Dataset~\cite{jaeger2014two} & Medical Image Diagnosis & 2 & 662 & \starh & \starh & \stare & \stare \\
        BrainTumor Dataset~\cite{liu2023chatgpt} & Medical Image Diagnosis & 2 & 593 & \starh & \stare & \stare & \stare \\
        IDRID Dataset~\cite{porwal2018indian} & Medical Image Diagnosis & 5 & 516 & \starh & \starh & \stare & \stare \\
        ISIC Dataset~\cite{codella2018skin} & Medical Image Diagnosis & 7 & 11720 & \stare & \starh & \stare & \stare \\

        HMDB-51~\cite{kuehne2011hmdb} & Action Recognition & 51 & 6,766 & \starh & \stare & \starh & \stare \\
        Kinetics-600~\cite{carreira2018short} & Action Recognition & 600 & 480,000 & \starh & \stare & \starh & \stare \\
        ActivityNet~\cite{caba2015activitynet} & Action Recognition \& Action Localization & 200 & 19,994 & \starh & \stare & \starh & \stare \\
        THUMOS14~\cite{idrees2017thumos} & Action Localization & 20 & 212 & \stare & \stare & \starh & \stare \\

        \bottomrule
        \end{tabular}
    }
\end{table*}

\subsection{Object Classification}
Object classification serves as a fundamental task for evaluating VLMs, where the objective is to assign a test object image to one of the candidate category names.
In the context of unsupervised adaptation with VLMs, research efforts primarily focus on two aspects: fine-grained generalization and robustness to distribution shifts.
To assess fine-grained classification performance, commonly used benchmark datasets include Caltech101~\cite{fei2004learning}, OxfordPets~\cite{parkhi2012cats}, StanfordCars~\cite{krause20133d}, Flowers102~\cite{nilsback2008automated}, Food101~\cite{bossard2014food}, FGVCAircraft~\cite{maji2013fine}, SUN397~\cite{xiao2010sun}, DTD~\cite{cimpoi2014describing}, EuroSAT~\cite{helber2019eurosat} and UCF101~\cite{soomro2012ucf101}.
To evaluate robustness against distributional shifts, researchers \cite{shu2022test, ma2024swapprompt, roth2023waffling} often employ ImageNet~\cite{deng2009imagenet} along with its variants, such as ImageNet-V2~\cite{recht2019imagenet}, ImageNet-Sketch~\cite{wang2019learning}, ImageNet-A~\cite{hendrycks2021natural}, and ImageNet-R~\cite{hendrycks2021many}.
Additionally, several studies \cite{tanwisuth2023pouf, liang2024realistic} incorporate datasets traditionally used in domain adaptation to evaluate their methods, such as Office-Home~\cite{venkateswara2017deep}, Office~\cite{saenko2010adapting}, and DomainNet~\cite{peng2019moment}.

\subsection{Semantic Segmentation}
Semantic segmentation aims to assign a semantic label to each pixel in an image, playing a critical role in applications such as autonomous driving and medical image analysis.
Unsupervised segmentation methods based on VLMs primarily focus on general and fine-grained object segmentation benchmarks, including PASCAL VOC 2012~\cite{everingham2015pascal}, PASCAL Context~\cite{mottaghi2014role}, COCO Stuff~\cite{caesar2018coco}, ADE20K~\cite{zhou2019semantic}, and COCO-Object~\cite{lin2014microsoft}.
In addition, complex scene understanding datasets such as Cityscapes~\cite{cordts2016cityscapes} and KITTI-STEP~\cite{weber2021step} are often employed to evaluate the performance of unsupervised segmentation approaches.
To assess the ability to identify rare concepts, some methods \cite{shin2022reco} utilize FireNet~\cite{firenet}.
Moreover, researchers~\cite{zhou2022extract} explore robustness to corruptions~\cite{hendrycks2019benchmarking} to find out whether segmentation algorithms preserve the inherent robustness of VLMs.
Segmentation performance is commonly quantified using the mean intersection-over-union (mIoU) metric.


\subsection{Visual Reasoning}
Context-dependent visual reasoning aims to identify whether a test image contains a given concept, based on a small set of support images that include both positive and negative examples.
The Bongard-HOI~\cite{jiang2022bongard} is commonly employed to assess the capability of VLMs to abstract the concept of human-object interaction from a limited number of support examples and accurately classify test samples.

\subsection{Out-of-Distribution Detection}
OOD detection focuses on identifying whether a test sample belongs to an in-distribution (ID) dataset composed of candidate categories, which plays a vital role in safety-critical applications.
Based on the degree of similarity between the OOD and ID datasets, OOD detection can be categorized into three main types: far OOD, near OOD, and fine-grained OOD detection.
Far OOD detection deals with samples that are clearly distinct from the ID distribution. For instance, when datasets such as CIFAR-100~\cite{krizhevsky2009learning}, CUB-200-2011~\cite{wah2011caltech}, StanfordCars~\cite{krause20133d}, Food101~\cite{bossard2014food}, OxfordPets~\cite{parkhi2012cats}, and ImageNet~\cite{deng2009imagenet} are used as ID data, datasets like iNaturalist~\cite{van2018inaturalist}, SUN397~\cite{xiao2010sun}, Places~\cite{zhou2017places}, and DTD~\cite{cimpoi2014describing} serve as typical far OOD sources.
Near OOD detection addresses a more challenging setting where the OOD samples share visual similarities with the ID data. Common experimental setups include alternately using ImageNet-10 and ImageNet-20 as ID and OOD datasets, as well as employing ImageNet-O~\cite{hendrycks2021natural} and OpenImage-O~\cite{krasin2017openimages} as near OOD sets.
Fine-grained OOD detection targets subtle distribution shifts within similar categories. For example, datasets such as CUB-200-2011~\cite{wah2011caltech}, StanfordCars~\cite{krause20133d}, Food101~\cite{bossard2014food}, and OxfordPets~\cite{parkhi2012cats} can be split such that half of the classes are seen as ID data and the other half as OOD data.
Evaluation of OOD detection performance is typically conducted using FPR95 and AUROC metrics.

\subsection{Text-Image Retrieval}
Text-image retrieval is a fundamental task in vision-language research, where the goal is to retrieve relevant images based on textual queries, or vice versa.
The MS-COCO~\cite{lin2014microsoft} and Flickr30K~\cite{plummer2015flickr30k} datasets are among the most widely used benchmarks for evaluating performance in this domain.
Besides, several specialized datasets are commonly utilized to assess retrieval performance across different contexts, such as Fashion-Gen~\cite{rostamzadeh2018fashion} from the e-commerce domain, CUHK-PEDES~\cite{li2017person}, ICFG-PEDES~\cite{ding2021semantically} from the person re-identification domain, and Nocaps~\cite{agrawal2019nocaps} from the natural image domain.
Recall@K serves as the standard metric for assessing the performance of retrieval algorithms.

\subsection{Image Captioning}
Image captioning aims to generate descriptive textual summaries of visual content.
In the context of test-time adaptation, researchers~\cite{zhao2023test} evaluate the adaptability of the CLIP model across several benchmark datasets, including MS-COCO\cite{lin2014microsoft}, Flickr30K\cite{plummer2015flickr30k}, and NoCaps\cite{agrawal2019nocaps}.
These datasets provide diverse visual and textual contexts, enabling the assessment of how effectively the model can generate relevant captions when exposed to new domains without additional annotated supervision.
Captioning performance is evaluated using BLEU, CIDEr, SPICE, and RefCLIPScore metrics.

\subsection{Beyond Vanilla Object Images}

\textbf{Medical image diagnosis.}
Medical imaging represents a critical real-world application of VLMs in unsupervised learning settings.
Researchers \cite{liu2023chatgpt, rahman2025can} frequently utilize datasets such as the Guangzhou Dataset~\cite{kermany2018identifying}, Montgomery Dataset~\cite{jaeger2014two}, and Shenzhen Dataset~\cite{jaeger2014two}, which focus on chest X-ray diagnosis.
Moreover, VLM-based unsupervised methods \cite{liu2023chatgpt, rahman2025can} have also been applied to various other diagnostic tasks, including diabetic retinopathy~\cite{porwal2018indian}, brain tumor detection~\cite{liu2023chatgpt}, and skin lesion classification~\cite{codella2018skin}.

\textbf{Videos.}
Beyond static images, VLMs have also been explored in the context of video-based unsupervised learning.
Action recognition benchmarks such as HMDB-51~\cite{kuehne2011hmdb}, UCF-101~\cite{soomro2012ucf101}, Kinetics-600~\cite{carreira2018short}, and ActivityNet~\cite{caba2015activitynet} are commonly used to evaluate performance.
In addition, more complex tasks like temporal action localization, which involve both action classification and precise timestamp prediction, are addressed using datasets such as ActivityNet~\cite{caba2015activitynet} and THUMOS14~\cite{idrees2017thumos}.

\section{Research Challenges and Future Directions}
\label{sec:future}
Despite significant progress, unsupervised VLM adaptation remains an open and challenging problem. This section outlines key research directions, identifying gaps in the current literature and discussing potential avenues for advancing the field.

\subsection{Theoretical Analysis}
While existing research has largely focused on developing effective unsupervised learning methods, rigorous theoretical analyses are still lacking. Understanding the theoretical complexities of VLMs is crucial for developing more principled adaptation methods. Future research can bridge this gap by providing formal generalization guarantees and characterizing the joint embedding space to explain how cross-modal alignment emerges~\cite{liang2022mind}. 

\subsection{Open-world Scenarios} 
Most existing approaches operate under the closed-set assumption, which presumes identical label spaces across domains. However, in real-world applications, test samples often contain unknown classes, making it essential to detect and handle them effectively. While some recent studies~\cite{sreenivas2024effectiveness,liang2024realistic,cao2025noisy} have begun addressing the open-world scenario, this challenging yet practical setting remains underexplored. Further research is needed to develop robust open-world adaptation methods that can generalize across diverse domains while accurately identifying unseen categories. Techniques from out-of-distribution detection~\cite{dong2024multiood,li2024dpu,liu2025fm} could also be leveraged and adapted to facilitate unknown class detection.

\subsection{Adversarial Robustness} 
Although VLMs demonstrate strong generalization capabilities, they remain highly susceptible to adversarial attacks~\cite{mao2022understanding}.
Several recent studies~\cite{mao2022understanding, li2024one} have drawn inspiration from adversarial training techniques~\cite{madry2017towards} to enhance the robustness of VLMs.
However, these approaches typically rely on large amounts of labeled data, leading to substantial annotation costs.
Therefore, an important research direction is to explore robust optimization~\cite{schlarmann2024robust} and inference strategies~\cite{sheng2025r} under unsupervised settings, enabling VLMs to operate reliably in complex, real-world environments where adversarial threats are likely and labeled data is scarce.

\subsection{Privacy Considerations} 
Privacy and security considerations are increasingly critical for the adaptation of VLMs, particularly in sensitive domains such as autonomous driving~\cite{dong2022dr} and healthcare~\cite{qayyum2020secure}. During adaptation, models may process proprietary or personal data, raising concerns about data leakage and unauthorized access. Additionally, the adaptation process can expose models to adversarial attacks~\cite{madry2017towards} that exploit vulnerabilities during the update phase, potentially degrading performance or leading to harmful outcomes. To address these challenges, future research should focus on developing privacy-preserving adaptation techniques such as federated learning~\cite{bao2023adaptive}, which enable models to adapt effectively without directly accessing raw data. 

\subsection{Efficient Inference} 
The deployment of VLMs demands substantial computational resources for inference. A critical research challenge is to reduce their latency and memory footprint without sacrificing performance. Future work may adapt techniques like quantization~\cite{lee2021network}, pruning~\cite{liu2018rethinking}, and knowledge distillation~\cite{gou2021knowledge} for the unique cross-modal nature of these models. The central difficulty lies in compressing the model while preserving the delicate vision-language alignments learned during pre-training. Developing novel and efficient architectures is crucial for enabling real-time VLM applications on resource-constrained hardware and moving these powerful models from the cloud to the edge.

\subsection{More VLMs Beyond CLIP} 
While CLIP has become the de facto backbone for unsupervised learning of VLMs, relying solely on its contrastive framework limits architectural and objective diversity. Future research should investigate alternative base models—such as advanced training strategies~\cite{zhai2023sigmoid}, masked-image modeling with joint text encoders~\cite{kwon2022masked}, or generative vision-language transformers~\cite{xie2024show}—to uncover new inductive biases. 
Moreover, studying how different encoder-decoder pairings impact alignment and transferability will guide the selection of more versatile models. Broadening beyond CLIP will catalyze novel unsupervised paradigms and improve VLM robustness across tasks and domains.

\subsection{Extension to MLLMs}
Another promising research direction is to integrate TTA into MLLMs with test-time scaling~\cite{muennighoff2025s1,snell2024scaling}. TTA methods enable models to dynamically adjust to distribution shifts during inference, enhancing robustness without retraining. In parallel, test-time scaling techniques allocate additional computational resources at test time—allowing models to "think" longer or perform deeper reasoning on challenging or out-of-distribution inputs~\cite{chen2024expanding}. By merging these approaches, an MLLM could not only adapt its predictions based on the incoming data stream but also flexibly scale its inference compute based on sample difficulty. This synergy would offer a balanced trade-off between efficiency and accuracy, especially in real-world applications where both rapid response and high adaptability are critical.

\subsection{New Downstream Tasks} 
Although unsupervised learning of VLMs has been extensively studied in image classification and semantic segmentation tasks, its potential in other domains remains largely underexplored, including regression~\cite{nejjar2023dare}, generative models~\cite{yang2023one}, cross-modal retrieval~\cite{li2024test}, depth completion~\cite{park2024test}, misclassification detection~\cite{dong2025trust}, and image super-resolution~\cite{deng2023efficient}. Besides, the potential applications in other fields such as medicine~\cite{kalpelbe2025vision} and healthcare~\cite{zhang2024generalist} remain underexplored and warrant greater attention.

\subsection{Failure Mode and Negative Transfer} 
Despite the empirical success of many unsupervised adaptation methods for VLMs, few studies have systematically documented their failure modes or reported instances of negative transfer. For example, entropy minimization~\cite{shu2022test}, although widely used, can reinforce incorrect predictions when the model exhibits high uncertainty, leading to overconfident misclassifications or even mode collapse. Similarly, prompt generation via LLMs may introduce hallucinated or domain-inappropriate descriptions~\cite{rawte2023survey}, resulting in semantic misalignment with the visual content and degraded performance. In continual adaptation settings~\cite{wang2022continual}, the accumulation of erroneous pseudo-labels over time can distort the feature space and destabilize the adaptation process.
To advance the field, future research should place greater emphasis on robustness analysis, including the development of metrics to detect adaptation failures and best practices for identifying and reporting instability. Furthermore, sharing negative results or counterexamples can play a critical role in uncovering systematic weaknesses and guiding the design of more resilient and reliable adaptation pipelines.


\section{Conclusion}
\label{sec:con}
In this survey, we have presented a comprehensive and structured overview of the rapidly advancing field of unsupervised vision-language model adaptation. Addressing a notable gap in existing literature, we introduced a novel taxonomy that classifies methods based on the availability of unlabeled visual data, a crucial factor for real-world deployment. By delineating the field into four distinct settings—data-free transfer, unsupervised domain transfer, episodic test-time adaptation, and online test-time adaptation—we provided a systematic framework for understanding the unique challenges and assumptions inherent to each scenario. Within this structure, we analyzed core methodologies and reviewed representative benchmarks, offering a holistic perspective on the state of the art. 
Finally, we identified several key challenges and directions for future research, including the development of theoretical analysis, the handling of open-world scenarios and privacy considerations, and further exploration of new downstream tasks and application fields.
This survey will not only serve as a valuable resource for practitioners seeking to navigate the landscape of unsupervised VLM adaptation but also stimulate further innovation by providing a clear basis for comparison and identifying promising directions for future research.


\ifCLASSOPTIONcaptionsoff
\newpage
\fi

\bibliographystyle{IEEEtran}
\bibliography{bib12, bib34}

\begin{IEEEbiographynophoto}{Hao Dong}
is a Ph.D. student at ETH Zürich, Zürich, Switzerland. He received the B.S. degree from Xi'an Jiaotong University, Xi'an, China, in 2020, and the M.S. degree from Aalto University, Espoo, Finland, in 2022. His research interest lies in multimodal learning and sensor fusion, including their applications in robotics, computer vision, and anomaly detection.
\end{IEEEbiographynophoto}

\vspace{-20pt}
\begin{IEEEbiographynophoto}{Lijun Sheng}
received the B.E. degree in Automation from the University of Science and Technology of China (USTC), in July 2020. 
He is now a Ph.D. candidate in the Department of Automation at the University of Science and Technology of China. His research interests include transfer learning, domain adaptation, and trustworthy AI.
\end{IEEEbiographynophoto}

\vspace{-20pt}
\begin{IEEEbiographynophoto}{Jian Liang}
received his B.E. degree in Electronic Information and Technology from Xi'an Jiaotong University in July 2013 and his Ph.D. degree in Pattern Recognition and Intelligent Systems from the National Laboratory of Pattern Recognition (NLPR), CASIA, in January 2019.
From June 2019 to April 2021, he worked as a research fellow at the National University of Singapore.
He is currently an associate professor at NLPR, CASIA. 
His research interests include transfer learning, pattern recognition, and computer vision.
\end{IEEEbiographynophoto}

\vspace{-20pt}
\begin{IEEEbiographynophoto}{Ran He}
received the BE degree in computer science from the Dalian University of Technology, in 2001, the MS degree in computer
science from the Dalian University of Technology, in 2004, and the PhD degree in pattern recognition and intelligent systems from CASIA, in 2009. 
Since September 2010, he joined NLPR, where he is currently a full professor. His research interests include information theoretic learning, pattern recognition, and computer vision. 
He serves as the editor board member of IEEE TPAMI, TIP, TIFS, TCSVT, and TBIOM, and serves on the program committee of several conferences. He is also a fellow of IEEE and IAPR.
\end{IEEEbiographynophoto}

\vspace{-20pt}
\begin{IEEEbiographynophoto}{Eleni Chatzi}
received the Diploma and M.Sc. degrees in civil engineering from the Department of Civil Engineering, National Technical University of Athens (NTUA), Athens, Greece, in 2004 and 2006, respectively, and the Ph.D. degree (with distinction) from the Department of Civil Engineering and Engineering Mechanics, Columbia University, New York, NY, USA, in 2010.
She is currently a Full Professor at the Chair of Structural Mechanics with the Department of Civil, Environmental and Geomatic Engineering, ETH Zürich, Zürich, Switzerland. Her research spans a broad range of topics, including applications on emerging sensor technologies and structural control, methods for curbing uncertainties in structural diagnostics and lifecycle assessment, as well as advanced schemes for nonlinear/nonstationary dynamics simulations. 
\end{IEEEbiographynophoto}

\vspace{-20pt}
\begin{IEEEbiographynophoto}{Olga Fink}
received the Diploma degree in industrial engineering from the Hamburg University of Technology, Hamburg, Germany, in 2008, and the Ph.D. degree from ETH Zürich, Zürich, Switzerland, in 2014.
She has been an Assistant Professor of Intelligent Maintenance and Operations Systems with EPFL, Lausanne, Switzerland, since March 2022. Before joining EPFL Faculty, she was an Assistant Professor of Intelligent Maintenance Systems with ETH Zürich from 2018 to 2022. From 2014 and 2018, she was heading the research group “Smart Maintenance” with the Zurich University of Applied Sciences, Winterthur, Switzerland. And, she is also a Research Affiliate with the Massachusetts Institute of Technology, Cambridge, CA, USA. Her research focuses on hybrid algorithms fusing physics-based models and deep learning algorithms, hybrid operational digital twins, transfer learning, self-supervised learning, deep reinforcement learning, and multiagent systems for intelligent maintenance and operations of infrastructure and complex assets.
\end{IEEEbiographynophoto}

\vfill

\end{document}